\documentclass[10pt,twocolumn,letterpaper]{article}

\usepackage{iccv}
\usepackage[ruled,vlined,linesnumbered]{algorithm2e}
\usepackage{amsmath}

\usepackage{multirow}
\usepackage{float}
\usepackage{pifont}
\usepackage{subcaption}
\usepackage{float}
\definecolor{iccvblue}{rgb}{0.21,0.49,0.74}
\usepackage[pagebackref,breaklinks,colorlinks,allcolors=iccvblue]{hyperref}
\usepackage{multirow}

\title{Unraveling the Effects of Synthetic Data on End-to-End Autonomous Driving}

\author{
Junhao Ge$^{1}$\footnotemark[1]
\quad Zuhong Liu$^{1}$\footnotemark[1]
\quad Longteng Fan$^{1}$\footnotemark[1]
\quad Yifan Jiang$^{1}$\footnotemark[1] \\
\quad Jiaqi Su$^{1}$
\quad Yiming Li$^{2}$
\quad Zhejun Zhang$^{3}$
\quad Siheng Chen$^{1,4}$\\
\normalsize{
$^{1}${Shanghai Jiao Tong University}
\quad $^{2}$New York University
\quad $^{3}$ETH Zurich
\quad $^{4}$Shanghai Artificial Intelligence Laboratory}\\
$^{1}${\small\texttt \{cancaries, albert\_liu, michaelfan, fiona.jiang, sjq2022, sihengc\}@sjtu.edu.cn} \\
$^{2}${\small\texttt \{yimingli\}@nyu.edu} \quad
$^{3}${\small\texttt \{zhejun.zhang\}@vision.ee.ethz.ch} \\
}

\newcommand{\acronym}{SceneCrafter}
\begin{document}
\twocolumn[{%
\maketitle
\vspace{-8mm}
\begin{figure}[H]
\vspace{-1.0mm}
\begin{center}
\hsize=\textwidth 
\includegraphics[width=0.9\textwidth]{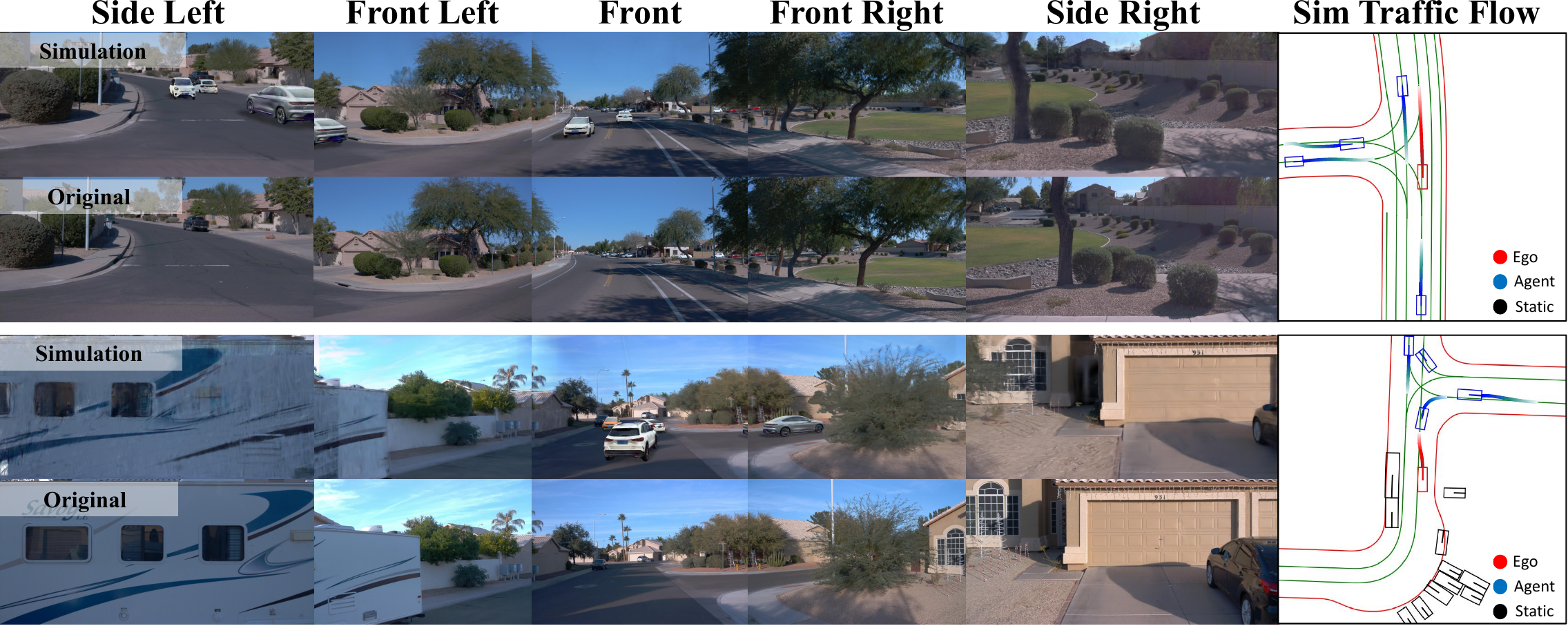}
\vspace{-2mm}
\caption{\acronym \ is a high-fidelity simulator capable of generating realistic synthetic driving data and providing an effective closed-loop evaluation for end-to-end autonomous driving models. Given real-world datasets, \acronym \ can reconstruct dynamic driving scenes and incorporate interactive traffic flows to generate novel, realistic, and consistent scenarios.}
\label{fig:teaser}
\end{center} 
\vspace{-3mm}
\end{figure}
}]

\maketitle
\begin{abstract}
End-to-end (E2E) autonomous driving (AD) models require diverse, high-quality data to perform well across various driving scenarios. However, collecting large-scale real-world data is expensive and time-consuming, making high-fidelity synthetic data essential for enhancing data diversity and model robustness. Existing driving simulators have significant limitations for synthetic data generation: game-engine-based simulators struggle to produce realistic sensor data, while NeRF-based and diffusion-based methods face efficiency challenges. Additionally, recent simulators designed for closed-loop evaluation provide limited interaction with other vehicles, failing to simulate complex real-world traffic dynamics. To address these issues, we introduce \acronym, a realistic, interactive, and efficient AD simulator based on 3D Gaussian Splatting (3DGS). \acronym\ not only efficiently generates realistic driving logs across diverse traffic scenarios but also enables robust closed-loop evaluation of end-to-end models. Experimental results demonstrate that \acronym\ serves as both a reliable evaluation platform and a efficient data generator that significantly improves end-to-end model generalization. Our code will be released at \url{https://github.com/cancaries/SceneCrafter}.
\end{abstract}    
\renewcommand{\thefootnote}{\fnsymbol{footnote}}
\footnotetext[1]{These authors contributed equally to this work.}
\section{Introduction}
\label{sec:intro}

Recently, autonomous driving (AD) has gradually progressed from modular paradigms~\cite{gonzalez2015review,kendall2019learning,chen2021data,xu2021autonomous} to end-to-end models~\cite{chitta2022transfuser,shao2023safety,chekroun2023gri,chitta2021neat,shao2023reasonnet,chen2022learning,hu2022st,hu2023planning,jiang2023vad,zheng2024genad}, which learn from sensor data and ego status to plan future motion. Most end-to-end models are fundamentally data-driven\cite{zheng2024preliminary}, where scenario diversity and vehicle trajectory distributions plays a decisive role in model effectiveness. However, gathering extensive, high quality and wide-coverage real data for driving scenarios is not only time-consuming but also costly. 

Given these constraints, generating high-fidelity, diverse synthetic data is essential for complementing real-world datasets, improving the robustness of end-to-end models. Nonetheless, current closed-loop simulators struggle to balance realism, efficiency, and interaction modeling, limiting high-quality synthetic data generation. Game engine-based approaches~\cite{dosovitskiy2017carla,yan2024oasim,qiao2024sumo,liu2024toend2ned,hanselmann2022king,li2022metadrive} provide extensive customization for scene editing; however, the sim-to-real gap in sensor realism limits their transferability. In contrast, diffusion-based methods~\cite{yang2024drivearena,ma2024unleashing,chang2025safe,yan2024drivingspherebuildinghighfidelity4d} enhance image realism but are constrained by high computational and time costs, as well as spatial-temporal inconsistencies in generated images. ~\cite{zhou2024hugsim} utilizes Gaussian Splatting for real-time image rendering but lacks reasonable interaction modeling for foreground objects. These limitations hinder current simulators from effectively generating synthetic data or fully evaluating end-to-end models. 

\begin{figure*}
 \vspace{-8.0mm}
  \centering
   \includegraphics[width=0.8\linewidth]{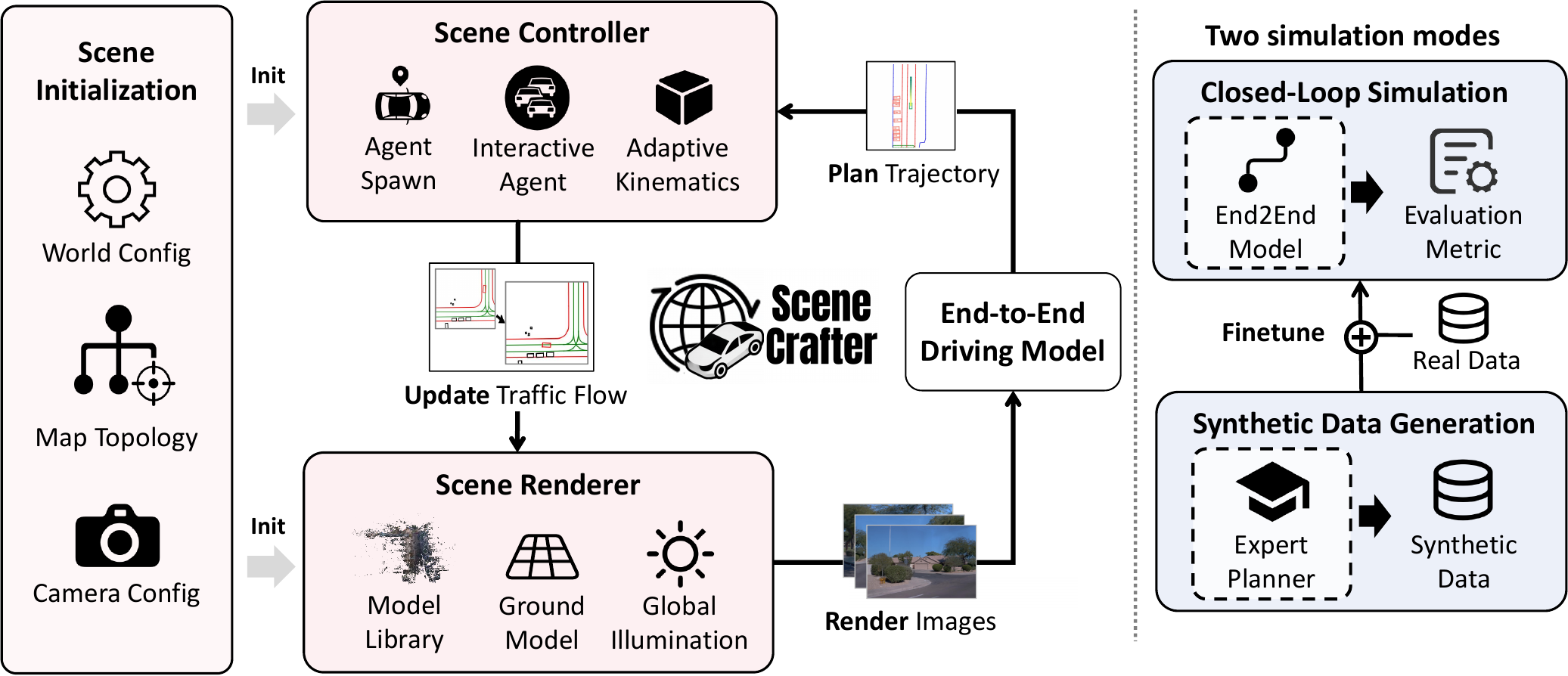}
   \vspace{-1.0mm}
   \caption{Overview of \acronym \ framework. Simulation is initialized with world configs, map topology and camera configs. Scene Controller updates interactive traffic flow, based on which Scene Renderer generates realistic driving scenes. End-to-End Driving Model plans future ego trajectory depending on simulation modes including synthetic data generation and closed-loop evaluation. }
   \vspace{-6mm}
   \label{fig:framework}
\end{figure*}

To address these challenges, we present \acronym, a unified simulation framework designed for high-quality synthetic data generation as well as closed-loop evaluation for end-to-end models.~\cref{fig:framework} shows the overview of \acronym . It consists of two key components: Scene Controller and Scene Renderer. Scene Controller initializes the simulation with real-world road map data and generates diverse, interactive traffic flows. Scene Renderer then integrates these foreground traffic agents into the reconstructed background, producing high-fidelity images that capture the evolving traffic scenario. Compared with existing methods, \acronym\ improves realism, interactivity and efficiency to better bridge the gap between simulation and real-world driving. ~\cref{tab:Comparison_w_current_works}
shows the comparison between our method and existing driving simulators. 

\begin{table}[t]
\footnotesize
\begin{center}
\centering
\setlength{\tabcolsep}{0.4mm}{
\begin{tabular}{cccccc}
\toprule 
\textbf{Name} & \begin{tabular}[c]{@{}c@{}}\textbf{Real}\\\textbf{Time}\end{tabular} & \begin{tabular}[c]{@{}c@{}}\textbf{ST}\\\textbf{Consistency}\end{tabular} & \begin{tabular}[c]{@{}c@{}}\textbf{Real}\\ \textbf{Images}\end{tabular} & \begin{tabular}[c]{@{}c@{}}\textbf{Interactive}\\ \textbf{Traffic Flow}\end{tabular} & \begin{tabular}[c]{@{}c@{}}\textbf{DataGen}\\ \textbf{For E2E}\end{tabular} \\
\midrule  
\textbf{CARLA}\cite{dosovitskiy2017carla} & \ding{51} & \ding{51} & \ding{55} & \ding{51} & \ding{51} \\
\textbf{SUMO}\cite{qiao2024sumo} & \ding{51} & \ding{51} & \ding{55} & \ding{51} & \ding{55} \\
\textbf{MetaDrive}\cite{li2022metadrive} & \ding{51} & \ding{51} & \ding{55} & \ding{51} & \ding{51} \\
\textbf{NeuroNCAP}\cite{ljungbergh2025neuroncap} & \ding{55} & \ding{51} & \ding{51} & \ding{55} & \ding{55} \\
\textbf{UniSim}\cite{yang2023unisim} & \ding{55} & \ding{51} & \ding{51} & \ding{55} & \ding{55} \\
\textbf{DriveArena}\cite{yang2024drivearena} & \ding{55} & \ding{55} & \ding{51} & \ding{51} & \ding{51} \\
\textbf{HugSim}\cite{zhou2024hugsim} & \ding{51} & \ding{51} & \ding{51} & \ding{55} & \ding{55} \\
\midrule  
\textbf{Ours} & \ding{51} & \ding{51} & \ding{51} & \ding{51} & \ding{51} \\ 
\bottomrule
\end{tabular}
}
\vspace{-3mm}
\caption{Comparison of existing and proposed simulation platforms for autonomous driving. \textbf{ST Consistency} represents Spatial-Temporal Consistency. \textbf{DataGen For E2E} represents Data Generation for end-to-end models.}
\vspace{-0.75mm}
\label{tab:Comparison_w_current_works}
\vspace{-10mm}
\end{center}
\end{table}

To achieve realistic simulation, we enhance ego agent dynamics, vehicle appearance and vehicle placement in images through several innovations. To capture real agents movement, our Scene Controller integrates an adaptive kinematic model trained on real-world IMU sensor data, ensuring realistic vehicle displacement and dynamics. Additionally, our Scene Renderer incorporates global illumination estimation to generate directional shadow Gaussian models, enhancing sensor image realism in dynamic scenes. To improve spatial modeling accuracy, we introduce an optimization-based ground height estimation algorithm that correct z-axis map offsets based on LiDAR points, ensuring precise vehicle positioning and rendering consistency. 

Unlike existing closed-loop simulators~\cite{zhou2024hugsim} that focus on agent-to-ego interactions, \acronym\ models bidirectional interactions between the ego vehicle and surrounding agents in Scene Controller. By dynamically updating agent behaviors through heuristic vehicle placement and actor models, it generates diverse and reasonable driving patterns to improve end-to-end model generalization.

Beides, \acronym\ leverages 3D Gaussian Splatting (3DGS) to model dynamic driving scenes. It achieves significantly higher efficiency and lower computational costs compared to NeRF-based~\cite{yang2023unisim,ljungbergh2025neuroncap}, or diffusion-based~\cite{yang2024drivearena,ma2024unleashing,chang2025safe,yan2024drivingspherebuildinghighfidelity4d} approaches, allowing the potential for large-scale synthetic data generation and evaluation. 

Our experiments demonstrate that the synthetic data generated by \acronym\ closely matches real-world sensor images and vehicle dynamics from the perspective of end-to-end models. We also assess the interactivity and efficiency of our method. Furthermore, we show that augmenting real-world datasets with our synthetic data enhances model generalization in interactive, closed-loop driving environments, achieving $18\%$ higher Route Completion rate.
In summary, our main contributions are as follows:
\begin{itemize}
    \item \textbf{Efficient and Reliable Synthetic Data Generation.}
    \acronym\ achieves superior efficiency and realism based on 3D Gaussian Splatting (3DGS).  This approach ensures efficient and reliable synthetic data production.
    \item \textbf{Interactive Closed-Loop Simulation Platform.} \acronym\ provides an interactive simulation environment with precise control over agent vehicles which allows comprehensive assessment of model robustness across diverse traffic interactions.
    \item \textbf{Enhanced End-to-End Models Generalization.}  With \acronym\ , we examine the influence of synthetic data on the training of the models. Our insights contribute to enhancing both model performance and generalization.
\end{itemize}

\section{Related Work}
\label{sec:related_work}

\noindent \textbf{End-to-End Model.} Recent attention in AD has shifted from modular paradigms~\cite{gonzalez2015review,kendall2019learning,chen2021data,xu2021autonomous} to end-to-end solutions~\cite{chitta2022transfuser,shao2023safety,chekroun2023gri,chitta2021neat,shao2023reasonnet,chen2022learning,hu2022st,hu2023planning,jiang2023vad,zheng2024genad}. These models integrate perception~\cite{li2022bevformer}, prediction~\cite{hu2021fiery}, and planning~\cite{renz2022plant} into a single framework, addressing issues such as information loss, error accumulation, and feature misalignment between modules~\cite{hu2023planning}.
 
Vision-based end-to-end models~\cite{hu2022st,hu2023planning,jiang2023vad,zheng2024genad} trained on real-world datasets attempt to extract more valuable information to enhance model performance. 
UniAD~\cite{hu2023planning} integrates information from preceding modules to achieve significant improvements in planning performance.
VAD~\cite{jiang2023vad} leverages the potential of vectorized scene representation, simplifying the input data and makes the model more efficient and scalable.
GenAD~\cite{zheng2024genad} enhances the multi-modality of ego vehicle trajectories using a generative approach, enabling more flexible motion planning. \cite{zheng2024preliminary} identifies scaling laws in imitation learning-based end-to-end autonomous driving models, demonstrating a power-law relationship between model performance and real training data volume. However, collecting large-scale real-world datasets is costly and time-consuming, highlighting the need for efficient methods to generate high-quality, diverse simulated data. Our work contributes to this direction by providing an approach to generating diverse, high-fidelity simulated data.

\noindent \textbf{Synthetic Data for Autonomous Driving. }
Data augmentation is a widely used technique for enhancing model performance across various autonomous driving tasks. Prior works~\cite{zhang2021exploringdataaugmentationmultimodality,lindstrom2024nerfs,tong20243d,chen2024s} have primarily focused on generating synthetic data to improve 3D perception. \cite{tong20243d} enhances 3D object detection by inserting 3D assets into a NeRF-reconstructed background using simple car replacement heuristics, while \cite{lindstrom2024nerfs} investigates the real-to-simulated domain gap by evaluating object detection and online mapping performance across different settings. 

However, limited research has explored the role of synthetic data in vision-based planning. \cite{goel2024syndiffadimprovingsemanticsegmentation} improves semantic segmentation and driving performance of end-to-end models in the CARLA simulator by generating data conditioned on semantic maps. To the best of our knowledge, this is the first study to investigate the impact of synthetic data on imitation learning-based end-to-end models trained on real-world driving data and evaluated in a photorealistic closed-loop environment.

\noindent \textbf{End-to-End Model Evaluation.} Most open-loop evaluations of end-to-end models compare the model's plan against GT records, such as L2 distance to GT trajectories or collision rates~\cite{caesar2020nuscenes,caesar2021nuplan,sun2020scalability}. However, this approach increasingly proves inadequate for assessing planning quality. Studies have identified distribution biases related to ego status~\cite{li2024ego,zhai2023rethinking}, leading models to learn shortcuts that fit training data but fail to generalize. In unseen scenarios, these models may struggle with adaptable planning or accurate trajectory prediction~\cite{bahari2022vehicle}. To overcome open-loop limitations, many simulators~\cite{dosovitskiy2017carla,yan2024oasim,qiao2024sumo,liu2024toend2ned,hanselmann2022king,li2022metadrive} provide robust closed-loop evaluation environments, though the sim-to-real gap hinders real-world transferability~\cite{sanders2016introduction,li2024scenarionet}. Recent NeRF-based~\cite{yang2023unisim,ljungbergh2025neuroncap} and diffusion-based~\cite{yang2023bevcontrol,swerdlow2024street,wen2024panacea,yang2024drivearena,ma2024unleashing,chang2025safe} methods generate realistic sensor data for closed-loop evaluation but suffer from high computational and time costs for large-scale testing. HugSim~\cite{zhou2024hugsim} reconstructs scenes using 3DGS but lacks reasonable and ego-to-agent interaction. In contrast, our work ensures realistic, interactive and efficient simulation.
\section{\acronym\ Framework}
\label{sec:system_overview}
\acronym\ offers a unified simulation framework supporting both synthetic data generation and closed-loop evaluation for end-to-end autonomous driving models. It consists of two components: Scene Controller and Scene Renderer. Scene Controller manages agents in the environment by dynamically updating their states and behaviors at each simulation step. Scene Renderer generates new driving scenes based on updates from Scene Controller and Driving Planner $\pi$, ensuring realistic and interactive environments. 

\subsection{Scene Controller}
\label{scene_controller}
Scene Controller dynamically updates the states and current behavior of ego based on their past states, ego vehicle's states, planned routes and map topology. It leverages control factors for different behavior modes, ensuring smooth and natural interactions in complex driving environments. 

\begin{figure}
  \centering
   \includegraphics[width=0.75\linewidth]{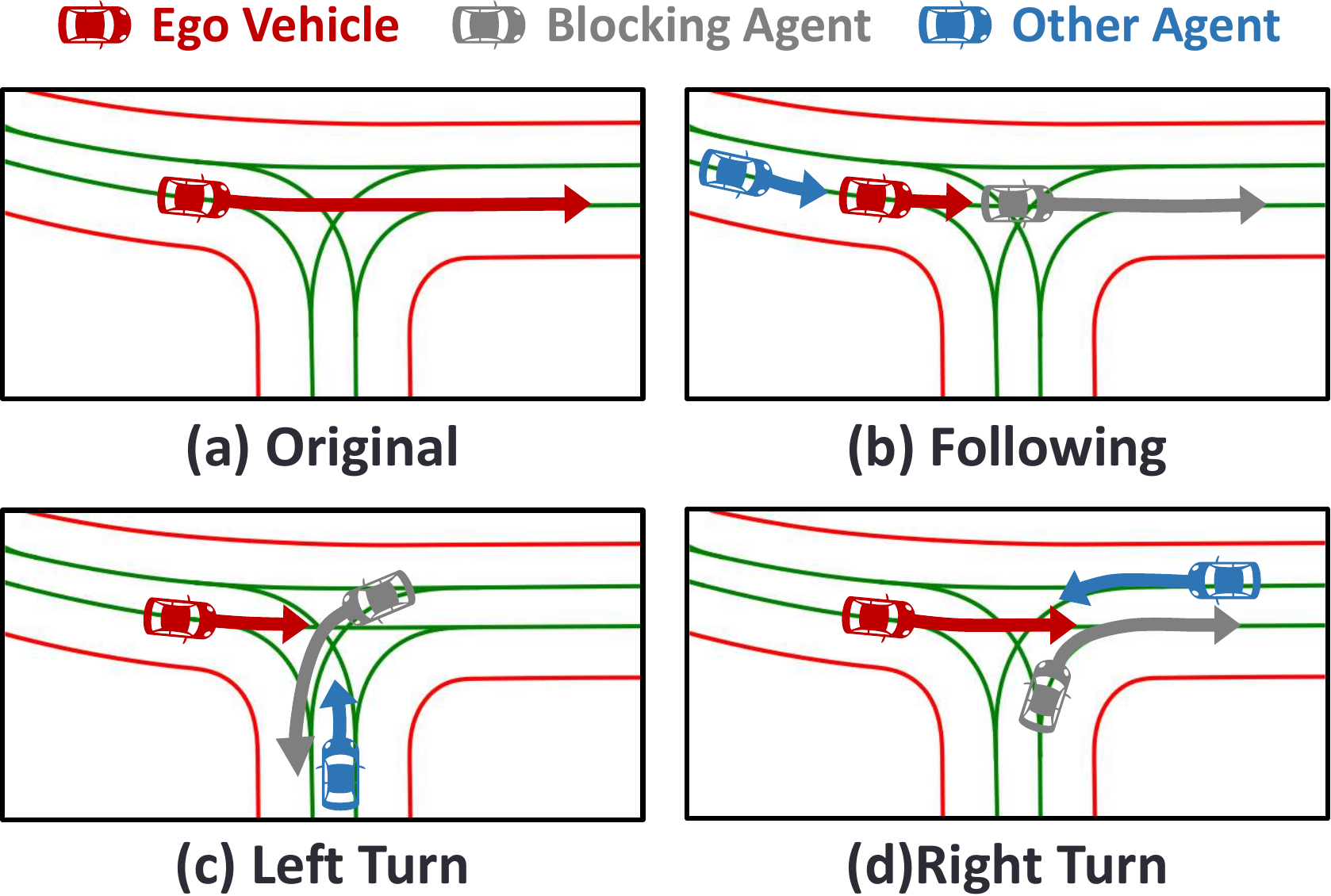}
   \vspace{-2.5mm}
   \caption{(a) illustrates the original scenario, where no vehicles are present except ego vehicle, which simply maintains a constant speed. In contrast, (b), (c), and (d) demonstrate how ego vehicle's behavior is altered due to the influence of the blocking agent, while ego vehicle impact the surrounding agents. Example video is given in supplementary material.}
   \label{fig:traffic_flow_example}
   \vspace{-8mm}
\end{figure}

\noindent \textbf{Heuristic Agent Spawn.}
The entire scene is initialized based on a configurable world setup $W$, including vehicle max spawn number, behavior modes and types. Scene Controller leverages HD maps derived directly from real-world datasets. By utilizing labeled information within HD maps ~\cite{sun2020scalability}, we build a map topological representation $\mathcal{M}$, which enables both precise navigation and localization.

With map topology, we can spawn $N$ agent vehicles by following two types of approaches: route-based and trigger-based. Agents are initialized with states $(s_{j})^{0}_{j=1:N}$ and behaviors $(b_{j})^{0}_{j=1:N}$. 
\begin{itemize}
\item \textbf{Route-based.} Spawn points with a higher likelihood of interaction with ego vehicle are prioritized as initialization conditions for agents. Then, we generate diverse global trajectories for agent vehicles, ensuring that the routes closely resemble real-world driving patterns.
\item \textbf{Trigger-based.} By selecting specific points along ego vehicle's trajectory as trigger points, we identify potential spawn points with map topology where agents could interact with ego vehicle. Once ego vehicle reaches a trigger point, the corresponding agent vehicles are activated to execute their trajectories. This approach can facilitate customized safety-critical test scenarios.
\end{itemize}
\cref{sec:agent_spawn_setting} explain these approaches in detail.

\noindent \textbf{Bi-interactive Agent Control.} For status and behavior control, we adopt an approach inspired by CARLA~\cite{dosovitskiy2017carla}, incorporating control factors such as speed limits and safe following distances derived from the original scenario. Agents determine driving behaviors $b^{t}_{j}$ based on their interaction with other agents at each time, which establishes a bidirectional interactive relation between ego and agent vehicles. ~\cref{fig:traffic_flow_example} shows some common interaction scenes. Beyond regular driving scenarios, we also introduce dangerous behavior modes for agent vehicles, which can trigger actions such as lane changes, aggressive overtakes, emergency stops, or disregarding safe distances when interacting with ego vehicle, providing potential for long-tail scenario evaluation.

\noindent \textbf{Adaptive Kinematic Model.} We propose an Adaptive Kinematic Model (AKM) to refine vehicle control signals, addressing the discrepancy between simulated and real ego dynamics during simulation, which can significantly impact planning \cite{li2024ego}. Given front length $l_f$ and rear length $l_r$, AKM updates vehicle states using ~\cref{eq:AKM}.
\begin{equation}
  \begin{aligned}[t]
            \beta_t &= \arctan\left( \frac{l_r}{l_f + l_r} \tan(\delta_t) \right), \quad v_{t+1}=v_{t}+a_t\Delta t,\\
        \varphi_{t+1} &= \varphi_{t} + \frac{v_t}{l_f} \sin(\beta_t)\Delta t, \quad v_u=\left(1-\boldsymbol{u_1}\right)v_t + \boldsymbol{u_1} v_{t+1},\\
        x_{t+1} &= x_t+v_u\cos(\varphi_t + \boldsymbol{u_2}\cdot\beta_{t})\Delta t, \\ 
        y_{t+1} &= y_t+v_u\sin(\varphi_t + \boldsymbol{u_2}\cdot\beta_{t})\Delta t.
  \end{aligned}
  \label{eq:AKM}
\end{equation}
where $\beta_t$ is the slip angle, $v_t$ is the vehicle velocity, $\varphi_t$ is the yaw angle, $\delta$ is the steering angle and $x_t$, $y_t $ represent the vehicle position. 
Compared with standard bicycle model, it incorporates two learnable parameters, $u_1$ and $u_2$, estimated from real-world IMU sensor data. These parameters help prevent unrealistic displacements by adjusting velocity and orientation dynamically. By ensuring smoother and more natural vehicle movement, this approach enhances the realism and continuity of the simulation environment.

\subsection{Scene Renderer}
\label{scene_renderer}
Scene Renderer generates high-fidelity driving scenes that reflect dynamic traffic flows with spatial-temporal consistency. It builds a Model Library using Gaussian Splatting for editable background and dynamic foreground models. We also propose ground height estimation and directional shadow rendering to enhance rendering realism. Figure \ref{fig:teaser} demonstrates example images from Scene Renderer.

\noindent \textbf{Gaussian-Splatting-Based Model Library.}  
Model library consists of two components: background model and foreground model, both are built based on Gaussian Splatting. For background, we train models by Street Gaussians~\cite{swerdlow2024street} with only static elements are retained to provide a stable background free from motion artifacts and allow for clean integration of dynamic foreground models. For foreground, due to the limited available angles of vehicles in the original scenario, we construct our virtual vehicle assets with BlenderNeRF~\cite{Raafat_BlenderNeRF_2024} and 3DRealCar~\cite{du20243drealcar} to build two foreground model libraries through training with 3DGS to meet the demands of various scenes. The explicit nature of Gaussian Splatting allows direct editing of the scene using models from the library. Detailed construction steps are provided in ~\cref{sec:gaussian_library}. 

\noindent \textbf{Ground Height Estimation.} To prevent artifacts from sinking into or floating above the ground in environments with large elevation variations, we propose learning a global ground model defined as $\hat{z}=f(x,y)$. Inspired by~\cite{chodosh2023reevaluatinglidarsceneflow}, we extract ground LiDAR points from each frame using RANSAC~\cite{10.1145/358669.358692} and fit a ground model for the entire scene, implemented as a three-layer MLP, to predict the z-axis offset during object placement. The loss is designed as follows to penalize less those LIDAR points above the ground:
\vspace{-1mm}
\begin{equation}
\mathcal{L}_{\text {height }}(\hat{z}, z)= \begin{cases}(\hat{z}-z)^2 & \hat{z}>z \\ \operatorname{Huber}(\hat{z}, z) & \hat{z} \leq z\end{cases}
\vspace{-1mm}
\end{equation}

\noindent \textbf{Directional Shadow Rendering.} To enhance the realism of foreground vehicles appearance in the scene, we generate vehicle shadow 3DGS models in multiple directions based on the Gaussian models from the foreground library. For each scene, we pre-estimate the global illumination based on the images from the original dataset as following:
\vspace{-1mm}
\begin{equation}
\mathbf{l} = \frac{ \begin{bmatrix} G_x(p^*), G_y(p^*), -1 \end{bmatrix}^\top }{ \sqrt{G_x^2(p^*) + G_y^2(p^*) + 1} }
\label{eq:light_dir}
\vspace{-1mm}
\end{equation}
where $p^* = \arg\max_p \| \nabla I(p) \|_2$ is the point with maximum gradient magnitude, 
$\nabla I(p) = [G_x, G_y]^\top$ denotes the Sobel gradient after Gaussian blur. During simulation, the global illumination and the current heading angle decide shadow model selection, which is combined with the original model for rendering. A visual comparison of foreground objects with shadow rendering and ground model placement is provided in ~\cref{fig:ground_comparison} in the supplementary.

\noindent \textbf{Composition Rendering.} Scene Renderer determines ego vehicle's viewpoint by camera configuration. Based on the agent vehicle poses provided by Scene Controller, foreground models are integrated into the background model. Ego vehicle's pose is then used to compute the camera's extrinsic parameters for rendering driving images. Utilizing composite Gaussian model with definite, precise control on spatial positions of foreground objects and overall appearance, our rendered images maintain spatial-temporal consistency, This ensures both the reliability of the generated data and the full controllability required for closed-loop evaluation. Detailed demonstration for spatial-temporal consistency can be found in~\cref{sec:spa-temp-consist}.
\subsection{Simulation Modes}
\acronym\ offers two simulation modes: synthetic data generation for realistic training and closed-loop simulation for interactive model evaluation. Algorithm \ref{algo_splatsim} outlines the simulation process for each mode.

\noindent \textbf{Synthetic Data Generation.}
To generate training datasets for end-to-end models, this mode employs an expert planner $\pi_{exp}$ to control the ego vehicle while dynamically updating surrounding agents.

Ego vehicle follows its original trajectory from the real dataset while adjusting its pace based on different driving scenarios to maintain consistent rendering quality. At each simulation step, our Scene Controller dynamically updates the status $s_j^{t}$ and behavior $b_j^{t}$ of surrounding agents.

The policy $\pi_{exp}$ generates control signals using privileged information $(s^{t}_{ego}, \left(s_{j}\right)^{t}_{j=1:N}, \mathcal{M})$ to ensure safe and natural driving behavior. The kinematic model $KM$ then applies these control signals to refine ego vehicle’s position and status for the Scene Renderer to generate new sensor observations. This cycle continues until either ego vehicle times out or reaches the target point.

\noindent \textbf{Closed-Loop Simulation.} 
To evaluate end-to-end models in realistic interactive environments, the closed-loop mode replaces the expert planner with a learned policy $\pi_{AD}$. Compared with data generation mode, the policy takes sensor inputs, control command and ego status as input to provide future action. The control command $c$ is generated from the original ego trajectory plan from real world data.

\begin{algorithm}
 \SetKwData{Up}{up}
 \SetKwFunction{Union}{Union}\SetKwFunction{SceneController}{SceneController}
 \SetKwFunction{SceneRenderer}{SceneRenderer}
 \SetKwInOut{Input}{Input}
\SetKwInOut{Initialize}{Initialize}
\SetKwInOut{Output}{Output}
\Input{World Setup $W$, Number of Agents $N$, Simulation Mode $M\in\{\text{SYN}, \text{CL}\}$, Driving Planner $\pi$, Kinematic Model \text{KM} } 
\Initialize{Map Topology $\mathcal{M}$,
Ego Initial State $s^{0}_{ego}$,
Agents Initial States $\left(s_{j}\right)^{0}_{j=1:N}$, \textbf{Agents Initial Behaviors} $\left(b_{j}\right)^{0}_{j=1:N}$, Sensor Images $(I_{k})^{0}_{j=1:C}$}
\Output{Driving Simulation Log $\tilde{\mathcal{D}}$ \\
$\{\left(s_{j}\right)^{1:T_{max}}_{j=1:N}, (I_{k})^{1:T_{max}}_{j=1:C},s^{1:T_{max}}_{ego},\mathcal{M}\}$}
\While{$t<T_{max}$}{
    \For{$j \leftarrow 1$ \KwTo $N$}{
        $s_{j}^{t+1}, \boldsymbol{b_{j}^{t+1}} \leftarrow$ \SceneController{$s^{t}_{\text{ego}}, \left(s_{j}\right)^{t}_{j=1:N}, b_{j}^{t}, \mathcal{M}$}
    }
    
    \eIf{$M = \text{SYN}$}{
        $\boldsymbol{{a_{\text{ego}}^{t+1} \leftarrow \pi_{\text{exp}}\left(s^{t}_{\text{ego}}, \left(s_{j+1}\right)^{t}_{j=1:N}, \mathcal{M}\right)}}$\tcp*{Synthetic data generation}
    }{
        $a_{\text{ego}}^{t+1} \leftarrow \pi_{\text{AD}}\left(s^{t}_{\text{ego}}, \left(I_{k}\right)^{t}_{k=1:C},c \right)$\tcp*{Closed-loop evaluation}
    }
    
    $s^{t+1}_{\text{ego}} \leftarrow \text{KM}\left(a^{t+1}_{\text{ego}}, s^{t}_{\text{ego}}\right)$\;
    $\left(I_{k}\right)^{t+1}_{k=1:C} \leftarrow$ \SceneRenderer{$s^{t+1}_{\text{ego}},\left(s_{j}\right)^{t+1}_{j=1:N}$}
    $t \leftarrow t+1$
}
\caption{\acronym\ Workflow}\label{algo_splatsim}
\end{algorithm}
\section{Experiment}
In Section~\ref{sec:exp_settings}, we explain our simulation setups for Scene Controller and Scene Renderer. In Section~\ref{sec:sim_fid_evaluation}, we share quantitative results that assess how well simulated data can replace real data, focusing on realism from the perspective of end-to-end autonomous driving models. Note that we do not compare our simulator with existing photorealistic simulators because they are unable to generate synthetic driving logs that include both sensor inputs and ego dynamics. Detailed justification are provided in ~\cref{sec:justify}. Section~\ref{sec:eval_interact} evaluates interactivity, while Section~\ref{sec:closed_loop_eval} assesses closed-loop performance, highlighting improvements in generalization and robustness. Additional runtime efficiency comparisons are provided in ~\cref{sec:runtime}.
\subsection{Simulation Settings}
\label{sec:exp_settings}
\noindent \textbf{Dataset.}
Our simulator is based on Waymo Open Dataset~\cite{sun2020scalability}, which is split according to its official configuration into 797 training scenes and 200 validation scenes. These scenes and their corresponding data samples are denoted as $(\mathcal{S}_{train}, \mathcal{D}_{train})$ for the training set and $(\mathcal{S}_{val}, \mathcal{D}_{val})$ for the validation set. For data generation, we select 60 representative scenes from the training set, covering both suburban and urban environments. These scenes include a variety of road types including straight roads and intersections, and span different weather conditions. The indices of these selected scenes are denoted as $\mathcal{S}_{recon} \subset \mathcal{S}_{train}$, while the rest indices are denoted as $\mathcal{S}^{c}_{recon}$. In addition, for closed-loop evaluation, we reconstruct 30 scenes from $\mathcal{S}_{val}$ to assess the generalization capability of end-to-end models.

\noindent \textbf{Scene Controller and Scene Renderer Setups.}
For Scene Controller, the updating frequency of vehicles' status in the scene is set to 10 Hz to ensure more precise control. The rendering frequency of Scene Renderer is set to 2 Hz following the input frequency of end-to-end models~\cite{hu2023planning,jiang2023vad,zheng2024genad} previously trained on nuScenes~\cite{caesar2020nuscenes}.

\noindent \textbf{End-to-End Models and Evaluation Metrics.}
We select VAD \cite{jiang2023vad} and its enhanced version, GenAD \cite{zheng2024genad}, as our end-to-end models due to their efficiency and effectiveness. Both methods shares the same auxiliary tasks: 3D Object Detection (3DOD), Motion Prediction and Online Map Segmentation (MapSeg). For open-loop evaluation, we assess planning performance using the L2 displacement error and collision rate, following the standard end-to-end benchmark~\cite{jiang2023vad,zheng2024genad}. In closed-loop evaluation, we adopt key metrics from Carla \cite{dosovitskiy2017carla}, including route completion (RC), vehicle collision rate (VCR), and layout collision rate (LCR). Detailed formulas are provided in \cref{sec:metrics}.

\subsection{Simulation Realism Evaluation}
To assess the reliability of \acronym , we examine the real-simulation gap between real data and generated data from the perspective of end-to-end models in an open-loop manner to show that \acronym\  can achieve real kinematic controls and high rendering quality for synthetic data.
\label{sec:sim_fid_evaluation}

\subsubsection{Data Augmentation with Synthetic Data }
To demonstrate the complementary benefits of our synthetic data, we train VAD with a mix of real-world and simulated data in an incremental learning setup. This enables us to assess whether the model surpasses one trained solely on real data, validating the fidelity of our synthetic data. Specifically, we construct two real-world training sets: $\mathcal{D}^{c}_{recon}$ and driving logs from 200 sampled scenes in $\mathcal{S}^{c}_{recon}$, forming a smaller-scale real training set $\mathcal{D}_{200}$ . We generate one driving log per scene from $\mathcal{S}_{recon}$ to create the synthetic training set $\tilde{\mathcal{D}}_{train}$ with around 2,000 samples. End-to-end models are evaluated on all samples $\mathcal{D}_{val}$ from $\mathcal{S}_{val}$.

\cref{tab:openloop} presents the quantitative results. The inclusion of synthetic data significantly improves planning performance when the real training set is small, demonstrating its potential to supplement real-world data. However, its impact is relatively marginal when the original training set is large. We attribute this to the fact that the Waymo validation set primarily consists of normal driving scenarios, such as cruising straightly in a scenario with fewer interaction with other agents, which are already well covered by large-scale real-world training data. This limitation of real-world open-loop evaluation benchmarks is also discussed in \cite{li2024ego}.
\begin{table}
\vspace{-2mm}
\footnotesize
\begin{center}
\centering
\setlength{\tabcolsep}{0.7mm}{
\begin{tabular}{l|cc|cccc|cccc}
\toprule
\multirow{2}{*}{\textbf{Setting}} & \multicolumn{2}{c|}{\textbf{Volume}} & \multicolumn{4}{c}{\textbf{Planning L2 (m) $\downarrow$}} & \multicolumn{4}{c}{\textbf{CR $\downarrow$}} \\
\cmidrule(lr){2-3} \cmidrule(lr){4-7} \cmidrule(lr){8-11}
 & \textbf{Real} & \textbf{Sim} & \textbf{1s} & \textbf{2s} & \textbf{3s} & \textbf{Avg} & \textbf{1s} & \textbf{2s} & \textbf{3s} & \textbf{Avg} \\
\midrule
  $\mathcal{D}_{200}$ & 8k & - & 0.61 & 1.31 & 2.24 & 1.39 & 0.34 & 0.36 & 0.86 & 0.52 \\
  $\mathcal{D}_{200}$ + $\tilde{\mathcal{D}}_{tr}$ & 8k & 2k & \textbf{0.48} & \textbf{1.09} & \textbf{1.99} & \textbf{1.18} & \textbf{0.28} & \textbf{0.32} & \textbf{0.75} & \textbf{0.45} \\
  \midrule
  $\mathcal{D}^{c}_{recon}$ & 28k & - & 0.43 & 1.02 & 1.82 & 1.09 & 0.21 & 0.27 & \textbf{0.47} & 0.32 \\
  $\mathcal{D}^{c}_{recon}$ + $\tilde{\mathcal{D}}_{tr}$ & 28k & 2k & \textbf{0.38} & \textbf{0.94} & \textbf{1.79} & \textbf{1.04} & \textbf{0.08} & \textbf{0.14} & 0.53 & \textbf{0.25} \\
\bottomrule
\end{tabular}
}
\vspace{-2mm}
\caption{Performance comparison of data augmentation with synthetic samples $\tilde{\mathcal{D}}_{tr}$ using different real data amounts $\mathcal{D}_{200}$ and $\mathcal{D}^{c}_{recon}$. The inclusion of synthetic data improves planning performance on both settings.
}
\label{tab:openloop}
\vspace{-6mm}
\end{center}
\end{table}

\subsubsection{Inference with E2E Models as Proxy Evaluators}

\begin{table*}
\vspace{-8mm}
\footnotesize
\begin{center}
\centering
\setlength{\tabcolsep}{2.3mm}{
\begin{tabular}{ccccccccccc}
\toprule
\multirow{2}{*}{\textbf{Setting}} & \multirow{2}{*}{\textbf{Volume}} & \multicolumn{2}{c}{\textbf{Rendering Config}} & \multicolumn{3}{c}{\textbf{Prediction (Car)}} & \multicolumn{1}{c}{\textbf{3DOD}$\uparrow$} & \multicolumn{3}{c}{\textbf{MapSeg}$\uparrow$} \\
\cmidrule(lr){3-4} \cmidrule(lr){5-7} \cmidrule(lr){8-8} \cmidrule(lr){9-11}
& & Car Lib & Shadow & EPA $\uparrow$ & ADE $\downarrow$ & FDE $\downarrow$ & Car AP & AP 0.5 & AP 1.0 & AP 1.5 \\
\midrule
Real Data & 2k & - & - & 0.30 & 0.71 & 0.96 & 44.1 & 0.245 & 0.482 & 0.644 \\
\midrule
VirVehicle Rep. & 2k & Virtual & \ding{55} & \textbf{0.25} & \textbf{0.26} & 1.02 & 38.4 & \textbf{0.227} & \textbf{0.476} & \textbf{0.633} \\
3DRealCar Rep. & 2k & 3DRealCar & \ding{51} & \textbf{0.25} & 0.77 & \textbf{1.01} & \textbf{39.6} & 0.212 & 0.467 & 0.622 \\
\bottomrule
\end{tabular}
}
\vspace{-2mm}
\caption{Perception and prediction performance comparison with VAD trained on $\mathcal{D}^{c}_{recon}$ over real and synthetic data. \textbf{Car Lib} represents the foreground object source, and \textbf{Shadow} denotes whether shadow rendering is applied. \textbf{3DRealCar Rep.} achieves a higher detection score compared with \textbf{VirVehicle Rep.}, demonstrating the enhanced realism of our foreground rendering.}
\label{tab:waymo_inference_image}
\vspace{-1mm}
\end{center}
\end{table*}

\begin{table*}[t]
\vspace{-3mm}
\footnotesize
\begin{center}
\centering
\setlength{\tabcolsep}{2.3mm}{
\begin{tabular}{ccccccccccc}
\toprule
\multirow[c]{2}{*}{\textbf{Setting}} & \multirow[c]{2}{*}{\textbf{Volume}} & \multicolumn{3}{c}{\textbf{Planning L2 (m)}$\downarrow$} & \multicolumn{3}{c}{\textbf{Prediction (Car)}} & \multicolumn{3}{c}{\textbf{MapSeg}$\uparrow$} \\
\cmidrule(lr){3-5} \cmidrule(lr){6-8} \cmidrule(lr){9-11}
& & 1s & 2s & 3s & EPA $\uparrow$ & ADE $\downarrow$ & FDE $\downarrow$ & AP 0.5 & AP 1.0 & AP 1.5 \\
\midrule
Real Data & 2k & 0.39 & 0.94 & 1.72 & 0.30 & 0.71 & 0.96 & 0.245 & 0.482 & 0.644 \\
\midrule
Sim Data w/ BM & 2k & 0.47 & 1.24 & 2.43 & 0.27 & 0.80 & 1.10 & 0.170 & 0.435 & 0.574 \\
\textbf{Sim Data w/ AKM (Ours)} & 2k & \textbf{0.38} & \textbf{1.04} & \textbf{2.06} & \textbf{0.28} & \textbf{0.77} & \textbf{0.92} & \textbf{0.185} & \textbf{0.453} & \textbf{0.590} \\
\bottomrule
\end{tabular}
}
\vspace{-2mm}
\caption{Inference results on Waymo reconstructed scenes with VAD trained on $\mathcal{D}^{c}_{recon}$. Our Adaptive Kinematic Model (AKM) produces ego trajectories closer to planning results than the Bicycle Model (BM), leading to improved prediction and perception performance. }
\label{tab:waymo_inference_kinematic}
\vspace{-6mm}
\end{center}
\end{table*}

\begin{figure}
  \centering
   \includegraphics[width=0.95\linewidth]{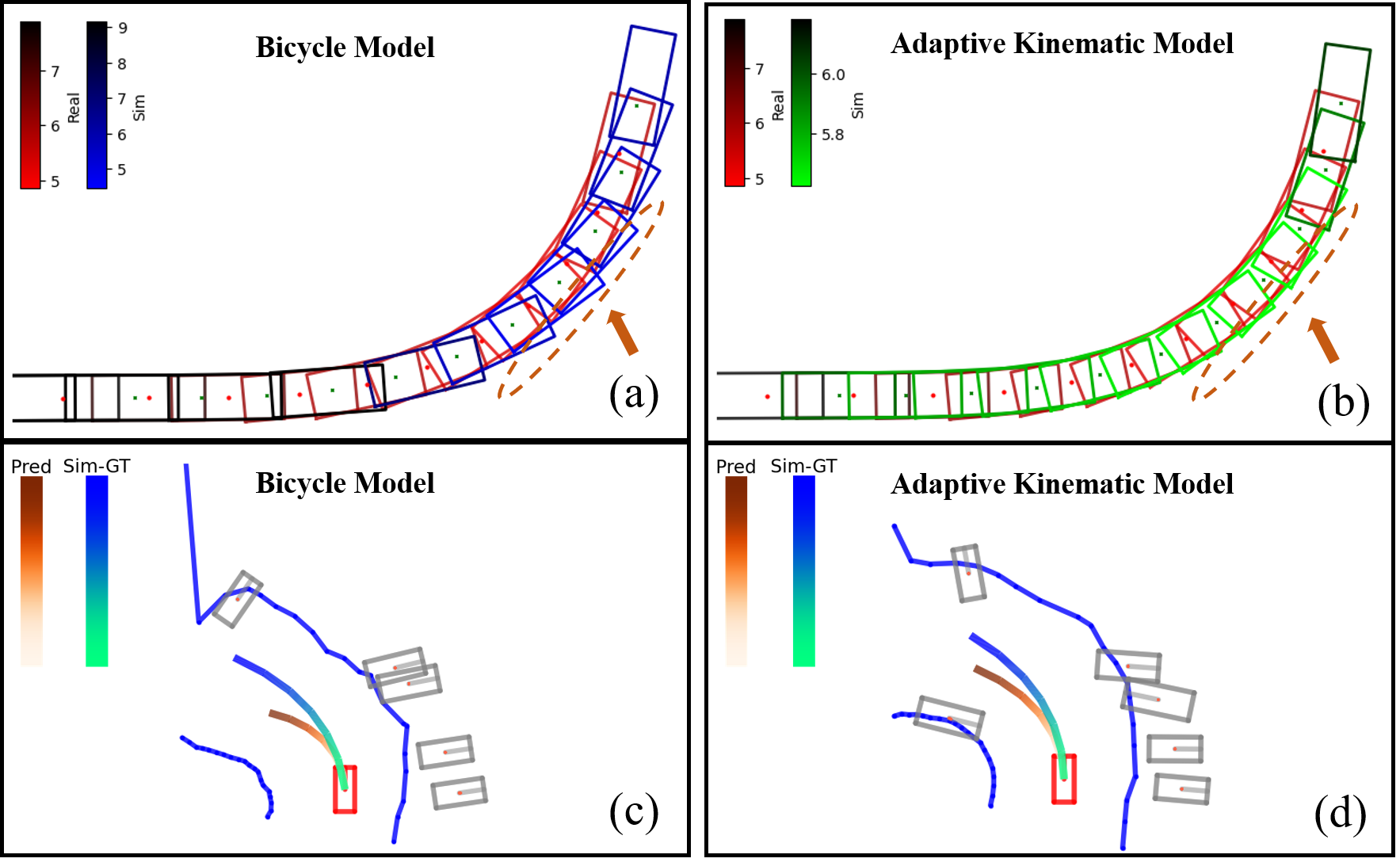}
   \caption{(a) and (b) show that ego trajectories generated by AKM are smoother and more realistic than those by BM. (c) and (d) illustrate that end-to-end planning result deviates more from the simulated GT trajectory with AKM than with BM, which indicates AKM aligns better with real world dynamics. The color bar indicates vehicle velocity norms.}
   \label{fig:kinematic_fidelity}
   \vspace{-6mm}
\end{figure}

Unlike most related works that primarily focus on image quality comparison for realism validation, we approach the problem by using VAD trained on real-world samples $\mathcal{D}^{c}_{recon}$ from scenes $\mathcal{S}^{c}_{recon}$ as a zero-shot proxy evaluator. This model is assumed to exhibit certain perception ability and reasonable coherence with the real world ground truth trajectory after large-scale training. We then evaluate the model on both real data $\mathcal{D}_{recon}$ and synthetic data $\tilde{\mathcal{D}}_{recon}$ generated from scenes $\mathcal{S}_{recon}$,  comparing their performance on all autonomous driving tasks and assessing the output trajectory of end-to-end models with the generated trajectory annotations by the simulator. This comparison aims to investigate the realism of the generated data from both the sensor input and vehicle kinematics perspectives. Specifically, we consider the following settings:
\begin{itemize}
    \item \textbf{Real Data:} This setting uses real-world driving logs from the scenes $\mathcal{S}_{recon}$ in the Waymo dataset. It includes actual sensor images and ego IMU data as inputs for VAD, along with annotated labels for other agents.
    \item \textbf{VirVehicle Rep.:} This setting uses real ego IMU data and annotated agents' labels from scenes $\mathcal{S}_{recon}$, but replaces the real sensor images with rendered images generated by Scene Renderer. The images are generated using the original camera configurations, with real dynamic foreground objects replaced by assets from the virtual vehicle library \cite{Raafat_BlenderNeRF_2024}, without adding shadows.
    \item \textbf{3DRealCar Rep.:} This setting mirrors \textbf{VirVehicle Rep.} but replaces real foreground objects with 3DRealCar\cite{du20243drealcar} assets and adding directional shadow during rendering.
    \item \textbf{Sim Data w/ BM:} This setting generates synthetic data with traffic flow controlled by Scene Controller, using a vanilla Bicycle Model as the kinematic model. The number of agents is set to match the average number of dynamic foreground objects in the original scenes.
    \item \textbf{Sim Data w/ AKM:} This setting generates synthetic data similar to \textbf{Sim Data w/ BM}, but uses AKM as the kinematic model instead of the vanilla Bicycle Model.
\end{itemize}

\cref{tab:waymo_inference_image} presents the realism assessment of images generated by Scene Renderer. Notably, replacing all foreground objects with reconstructed 3D assets and rendered background model results in only a minor decline in perception performance (3DOD and MapSeg), demonstrating the high fidelity of our rendered images.  This performance gap is primarily due to an increase in false positive bounding boxes, which is less critical than false negatives. Additionally, foreground objects from 3DRealCar, which include added shadows, achieve higher detection scores compared to virtual vehicles, further validating their realism.

Regarding the kinematic model, ~\cref{tab:waymo_inference_kinematic} shows that the ego trajectory generated by the Adaptive Kinematic Model (AKM) aligns more closely with planning results than that generated by the simpler Bicycle Model. ~\cref{fig:kinematic_fidelity} further illustrates the differences in turning trajectory annotations compared to real trajectories. The ego trajectory produced by the Bicycle Model often results in unnatural vehicle orientations, whereas AKM corrects this issue through its learnable parameters, leading to more realistic ego motions.

\begin{figure*}
  \centering   \includegraphics[width=0.82\linewidth]{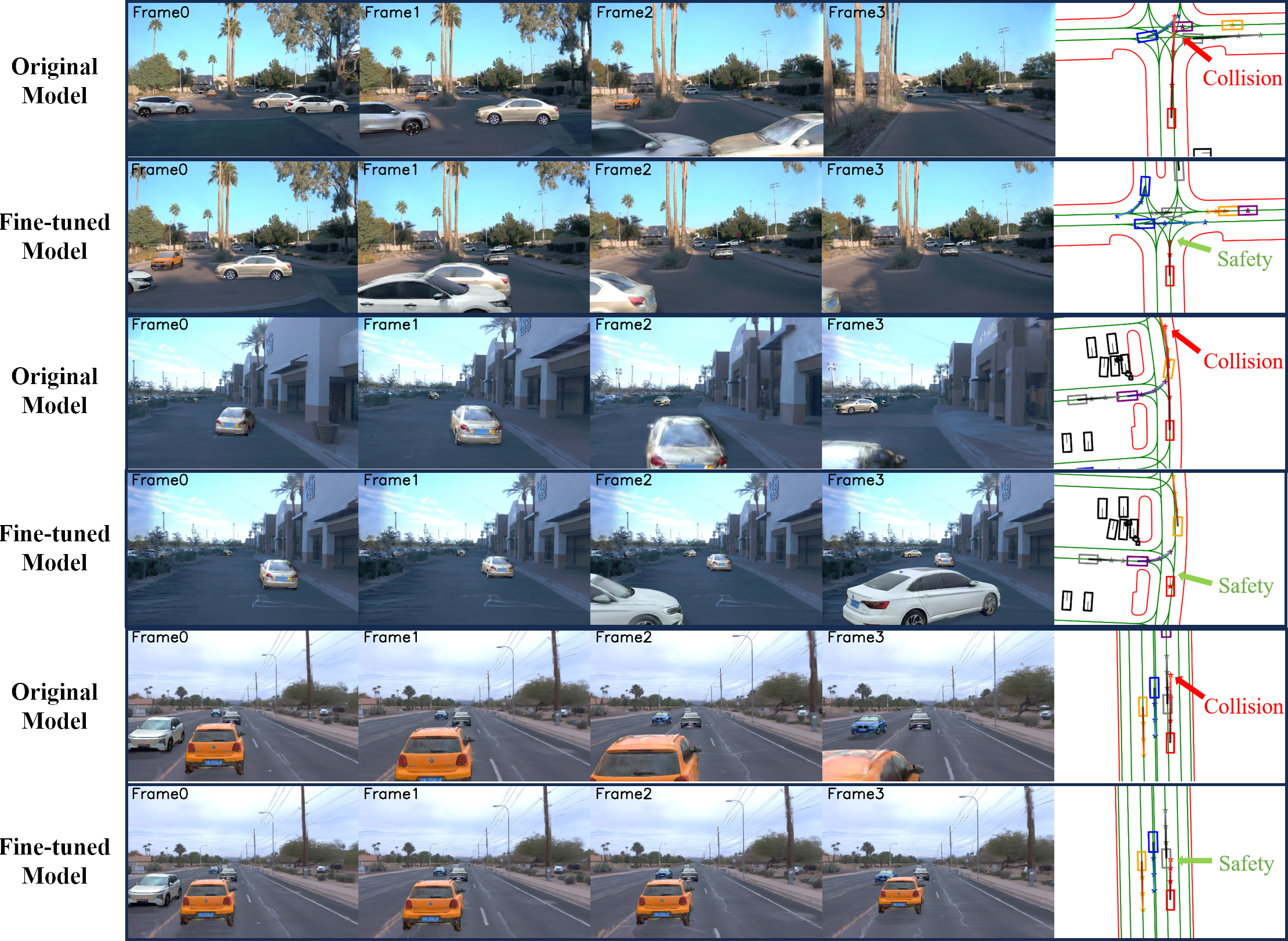}
  \vspace{-2.0mm}
   \caption{Qualitative comparison from front camera view of the original end-to-end models and those fine-tuned with our synthetic data. 
 The \textcolor{red}{Red} car represents ego vehicle controlled by $\pi_{AD}$. The \textcolor{red}{red} and \textcolor{green}{green} arrows respectively indicate the locations where the collision occurs or is avoided. Fine-tuned models offer more reasonable driving behaviors in complex scenarios while models solely trained on real data, showcasing enhanced generalization.}
\label{fig:failure_case}
\vspace{-5mm}
\end{figure*}

\subsection{Interactivity Evaluation}
\label{sec:eval_interact}
To assess our interactivity, we compare statistics including bidirectional interactions and ego speed alteration in scenes from $S_{recon}$ on our synthetic data and real data. Interaction Rate measures the proportion of scenes containing either ego-to-agent or agent-to-ego interactions across all scenes. Ego Speed Alteration quantifies how often the ego vehicle's speed tendency shifts (from acceleration to deceleration or vice versa) within the next $t=6$ time steps due to interactions with other vehicles.

\begin{figure}
  \centering
   \includegraphics[width=0.9\linewidth]{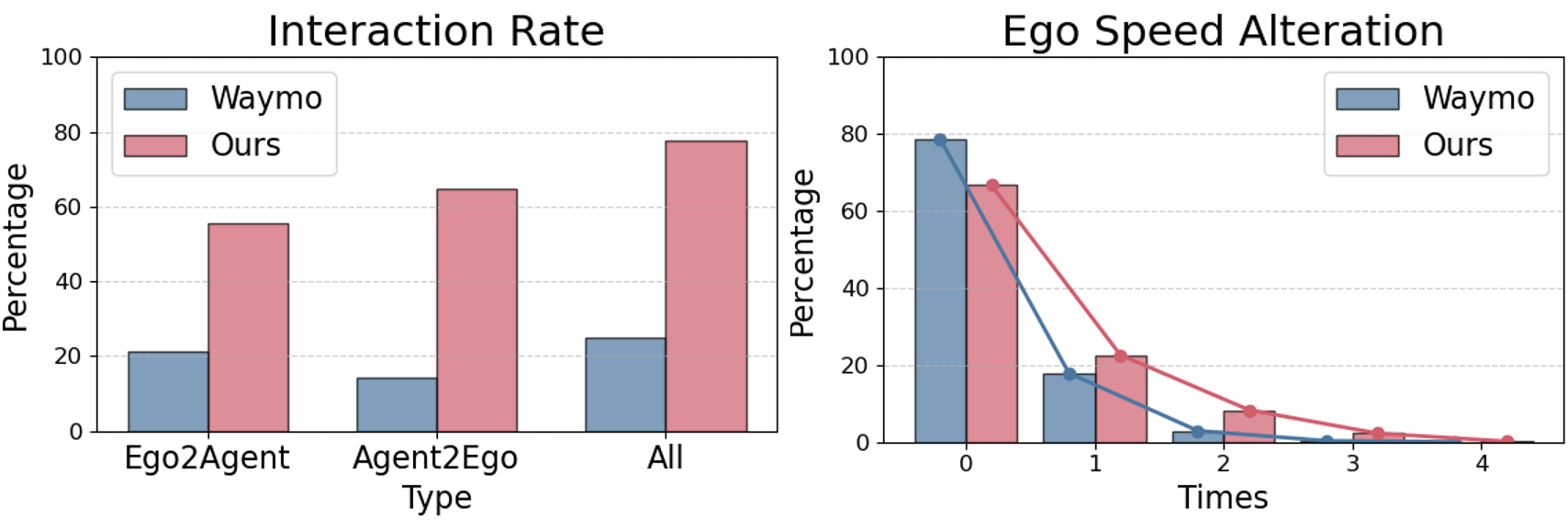}
   \vspace{-2mm}
   \caption{Our synthetic data shows a higher Interaction Rate and a longer-tail distribution of Ego Speed Alteration, reflecting greater interactivity and diversity in generated traffic flow.}
   \vspace{-5mm}
   \label{fig:interactivity}
\end{figure}

As shown in \cref{fig:interactivity}, our synthetic data exhibits a higher interaction rate compared to the original Waymo dataset. Furthermore, the Ego Speed Alteration metric reveals a longer-tail distribution than real data, indicating greater diversity in ego vehicle speed variations influenced by surrounding agents. This variability helps prevent models from learning flawed patterns, such as consistently accelerating or decelerating that are more prevalent in real-world data. These findings highlight the enhanced diversity and interactivity of our simulation, reinforcing its value as a complementary resource to real-world datasets. Detailed definitions of the Ego Speed Alteration and Interaction Rate metrics are provided in \cref{sec:interaction_judge}.

\subsection{Closed-Loop Evaluation}
\label{sec:closed_loop_eval}
With \acronym\ generating realistic sensor data, it also enables interactive closed-loop evaluation, providing a more comprehensive assessment of end-to-end model performance in real driving scenarios. To see how our data contributes to the enhancement of models, we generate three random trajectories per scene with more involved agents than in ~\cref{sec:sim_fid_evaluation} to construct a synthetic dataset for fine-tuning VAD and GenAD which are initially trained on $\mathcal{D}_{train}$. We initialize each scene with three different seeds and the ego vehicle at two starting positions.

\begin{table}
\footnotesize
\begin{center}
\centering
\setlength{\tabcolsep}{1.5mm}{
\begin{tabular}{l|cc|cccccc}
\toprule
\multirow{3}{*}{\textbf{Model}} & \multicolumn{2}{c|}{\textbf{Volume}} & \multicolumn{3}{c}{\textbf{All Test Scenes}} & \multicolumn{3}{c}{\textbf{Turn Scenes}} \\
\cmidrule(lr){2-3} \cmidrule(lr){4-6} \cmidrule(lr){7-9}
 & Real & Sim & \textbf{RC$\uparrow$} & \textbf{VC$\downarrow$} & \textbf{LCR$\downarrow$} & \textbf{RC$\uparrow$} & \textbf{VC$\downarrow$} & \textbf{LCR$\downarrow$} \\
\midrule
  \multirow{2}{*}{VAD} & 32k & - & 42.01 & 7.55 & 7.48 & 25.71 & 7.90 & 9.84 \\
   & 32k & 6k & \textbf{53.03} & \textbf{3.43} & \textbf{6.08} & \textbf{34.72} & \textbf{3.78} & \textbf{8.08} \\
  \midrule
  \multirow{2}{*}{GenAD} & 32k & - & 44.87 & 7.62 & 8.50 & 28.47 & 7.23 & 9.25 \\
   & 32k & 6k & \textbf{49.65} & \textbf{5.45} & \textbf{5.76} & \textbf{34.02} & \textbf{5.63} & \textbf{8.14} \\
\bottomrule
\end{tabular}
}
\vspace{-2mm}
\caption{Closed-loop evaluation performance comparison. Metrics include: \textbf{RC} (Route Completion), \textbf{VC} (Vehicle Collision Rate), and \textbf{LCR} (Layout Collision Rate). Models fine-tuned with synthetic data exhibit improved performance in closed-loop evaluation compared to models trained on real data alone.}
\vspace{-6mm}
\label{tab:close_loop2}
\end{center}
\end{table}

\noindent \textbf{Generalizable Capacity.} ~\cref{tab:close_loop2} shows the results of original and fine-tuned models evaluated on unseen scenes from $\mathcal{S}_{val}$. We see that for both VAD and GenAD , the fine-tuned versions achieve a much higher route completion with a lower collision rate with dynamic agents and static map elements. This improvement in closed-loop evaluation further proves the feasibility to improve AD models' generalization ability by simulated data.

\noindent \textbf{Qualitative Results.}
~\cref{fig:failure_case} shows qualitative comparison between original and fine-tuned end-to-end models. The fine-tuned models demonstrate improved performance in various complex driving scenarios, including intersection handling, highway car-following, and lane merging, highlighting their enhanced generalization capabilities.

\section{Conclusion}
\noindent\textbf{Summary.} In this paper, we introduce \acronym, a unified simulator designed for training and testing vision-based end-to-end autonomous driving models. By leveraging 3DGS, \acronym\ efficiently generates spatial-temporally consistent images, which is crucial for both data generation and closed-loop evaluation. To enhance realism, \acronym\ integrates an adaptive kinematic model and high-fidelity foreground rendering. Additionally, it creates an interactive and responsive environment using heuristic vehicle placement and actor models, enabling real-world-level interactions. Experimental results demonstrate the high fidelity of our framework and highlight its potential to improve the generalization capability of autonomous driving models in real-world scenarios.

\noindent\textbf{Limitation and future work.} Current types of traffic participants are limited. Future work will focus on increasing the agent diversity, such as cyclist and pedestrian, to further enrich traffic scenarios and make the interaction more challenging. Besides, We aim to enhance this by incorporating lightweight local diffusion models to address unseen scenarios, thereby expanding the rendering range and improving overall rendering quality.
{
    \small
    \bibliographystyle{ieeenat_fullname}
    \bibliography{main}
}
\clearpage
\appendix
\setcounter{page}{1}
\maketitlesupplementary

\section{Agent Spawn Setting Setting}
\label{sec:agent_spawn_setting}
\begin{figure}
  \centering
   \includegraphics[width=\linewidth]{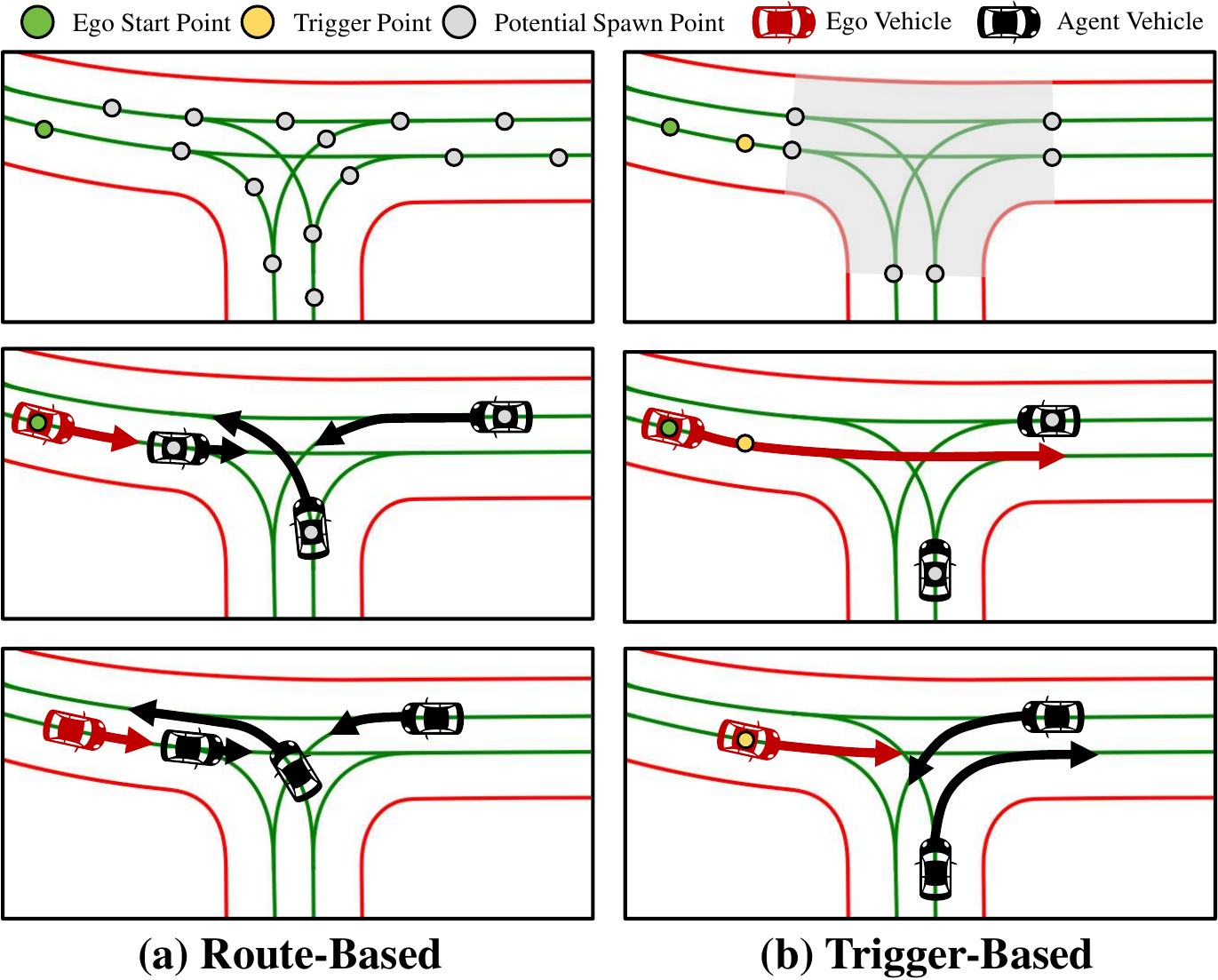}
   \vspace{-5mm}
   \caption{Illustration of agent spawn settings.}
   \label{fig:agent_spawn_settings}
   \vspace{-6mm}
\end{figure}

As mentioned in ~\cref{sec:system_overview}, we adopt two agent spawn methods: route-based and trigger-based. ~\cref{fig:agent_spawn_settings} illustrates these two spawn methods.

\noindent \textbf{Route-Based.} With the map topology, we first sample lane points at regular intervals as candidate spawn points. From these points, we generate routes and filter out those whose shorter than a predefined distance threshold. We then assess whether these routes interact with ego vehicle’s trajectory, forming a set of candidate agent routes. Each time an agent is placed, we remove candidate spawn points within a certain radius to control spawn density and ensure a reasonable scene distribution. Once the agent spawn is complete, all agent vehicles follow normal behavior control mechanisms. This approach can lead to complex and diverse interactions, enriching the generated simulation data.

\noindent \textbf{Trigger-Based.} We randomly select points along ego vehicle's trajectory as trigger points, with a higher preference for those located near intersections. Using the map topology, we locate the intersection closest to the route following the trigger point and identify all lane endpoints within the intersection as candidate spawn points for agent vehicles. For route generation, we select the most interactive route from road options such as Follow, TurnLeft, TurnRight, LaneChangeLeft, and LaneChangeRight, based on road topology and its interaction potential with ego vehicle. Unlike the route-based approach, spawned agents in the trigger-based method do not start moving immediately. Instead, they are activated and assigned an initial state only when the ego vehicle reaches the trigger point. Compared to the route-based method, the trigger-based approach ensures interaction and allows further expansion of possible events based on road topology.

\section{3D Gaussian Model Library}
\label{sec:gaussian_library}
\begin{figure*}
  \centering
   \includegraphics[width=0.9\linewidth]{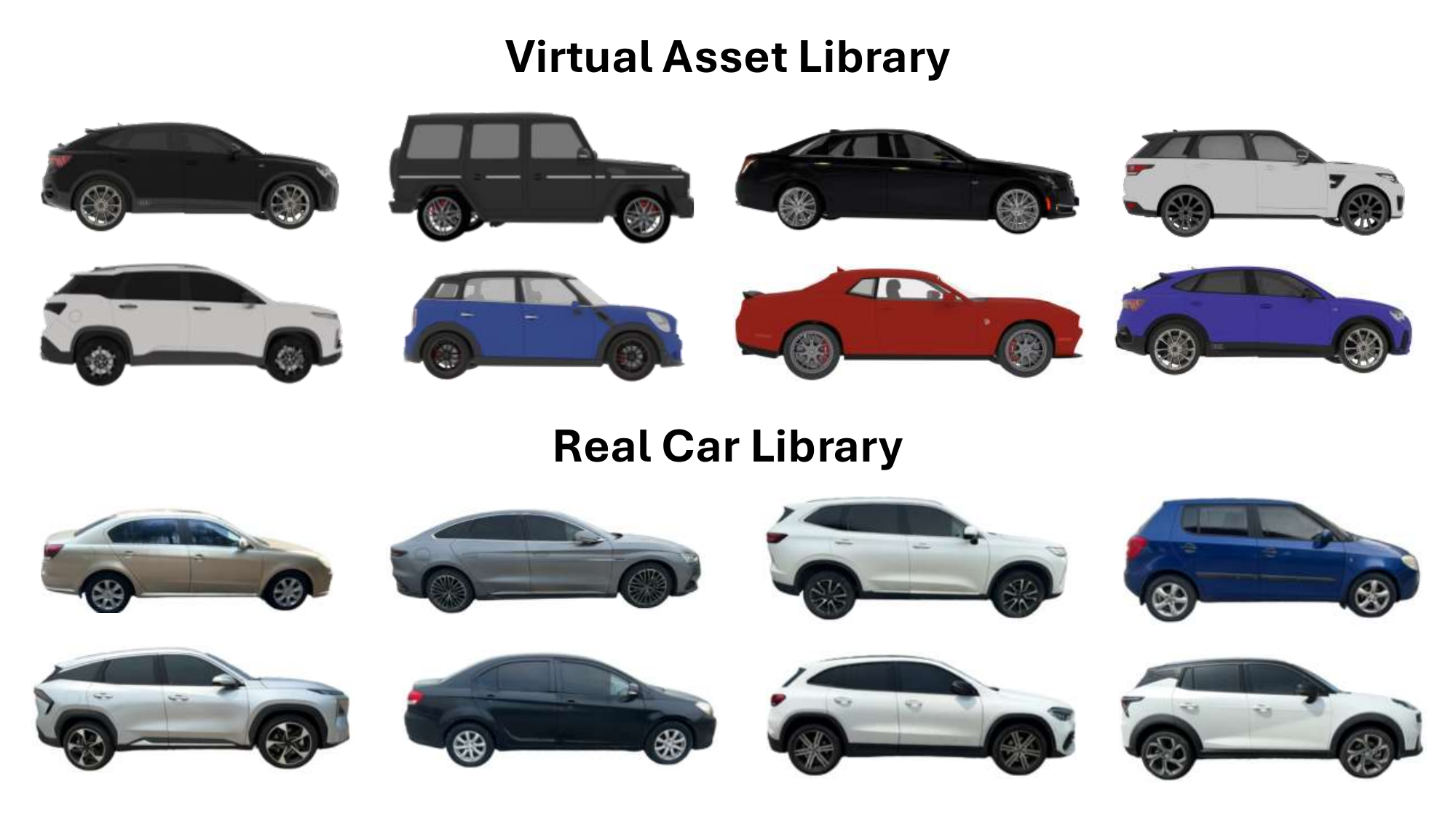}
   \vspace{-2mm}
   \caption{Overview of foreground gaussian model library.}
   \label{fig:library_foreground}
\end{figure*}

\begin{figure*}
  \centering
   \includegraphics[width=0.9\linewidth]{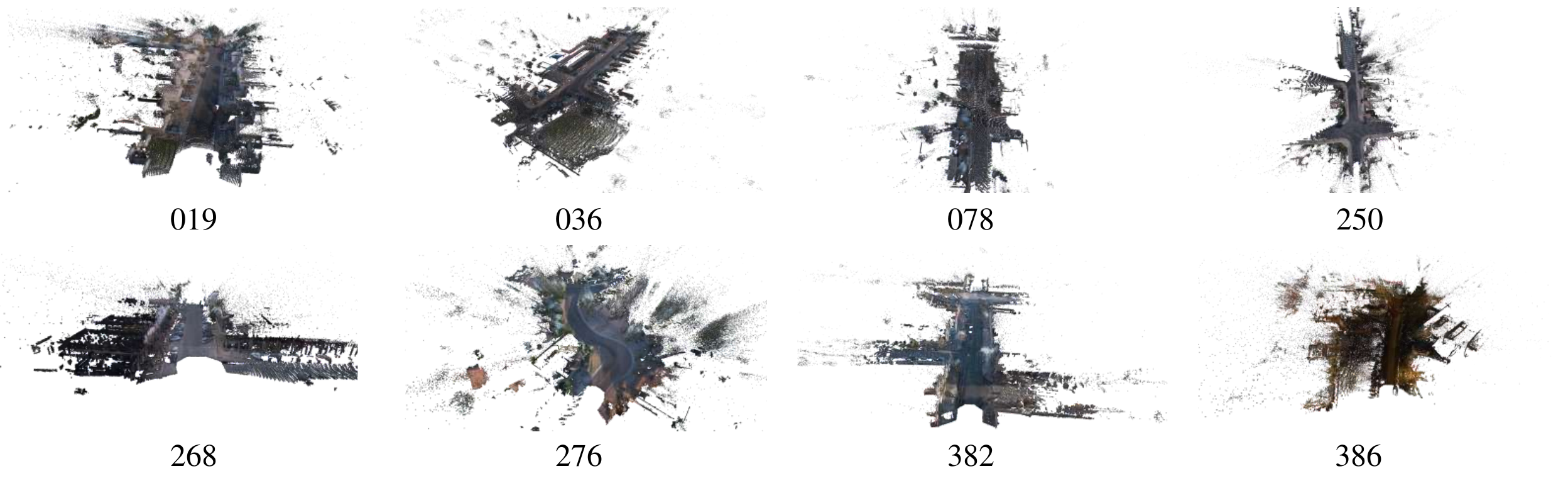}
   \vspace{-2mm}
   \caption{Overview of background gaussian model library.}
   \label{fig:library_background}
\end{figure*}

Figures \ref{fig:library_foreground} and \ref{fig:library_background} illustrate the foreground models and static models used in our experiments. During open-loop benchmark generation and closed-loop simulation, we select a specific background model from the background model library and register each agent vehicle using one of the foreground models from the foreground model library, which remains consistent throughout the process.

\subsection{Background Rendering Detail}
\noindent \textbf{Dataset Selection.} We select scenarios from Waymo Open Dataset\cite{sun2020scalability} and categorize them based on time and vehicle behavior. Temporally, the scenarios are divided into Day and Night. Behaviorally, they are divided into Straight and Turn depending on whether the vehicle executes a turning maneuver within the scenario. With these classifications, we can easily select suitable scenes to better generate targeted scenarios to address corner cases. 

\noindent \textbf{Model Training.} The training process for our background models follows Street Gaussians. Unlike Street Gaussians\cite{yan2024sg}, which focus on rendering images from the front three camera views, \acronym\ requires a wider field of view for both open-loop benchmark generation and closed-loop simulation. Therefore, we use images from all five cameras in Waymo Open Dataset to train background models. This ensures that even when adjusting parameters such as camera field of view and relative positioning during simulation, we can still produce high-quality rendered images.

\subsection{Foreground Rendering Detail}
Our foreground vehicles consist of two parts: one is the virtual assets generated using BlenderNeRF, and the other comes from 3DRealCar. During training, we observe that the real dataset had limitations due to the absence of images from certain perspectives, such as the top and bottom views, as well as reflection challenges. These limitations lead to trained models that do not generalize well in the simulation environment. To address this issue, we explore two approaches.

\begin{figure*}
  \centering
   \includegraphics[width=0.9\linewidth]{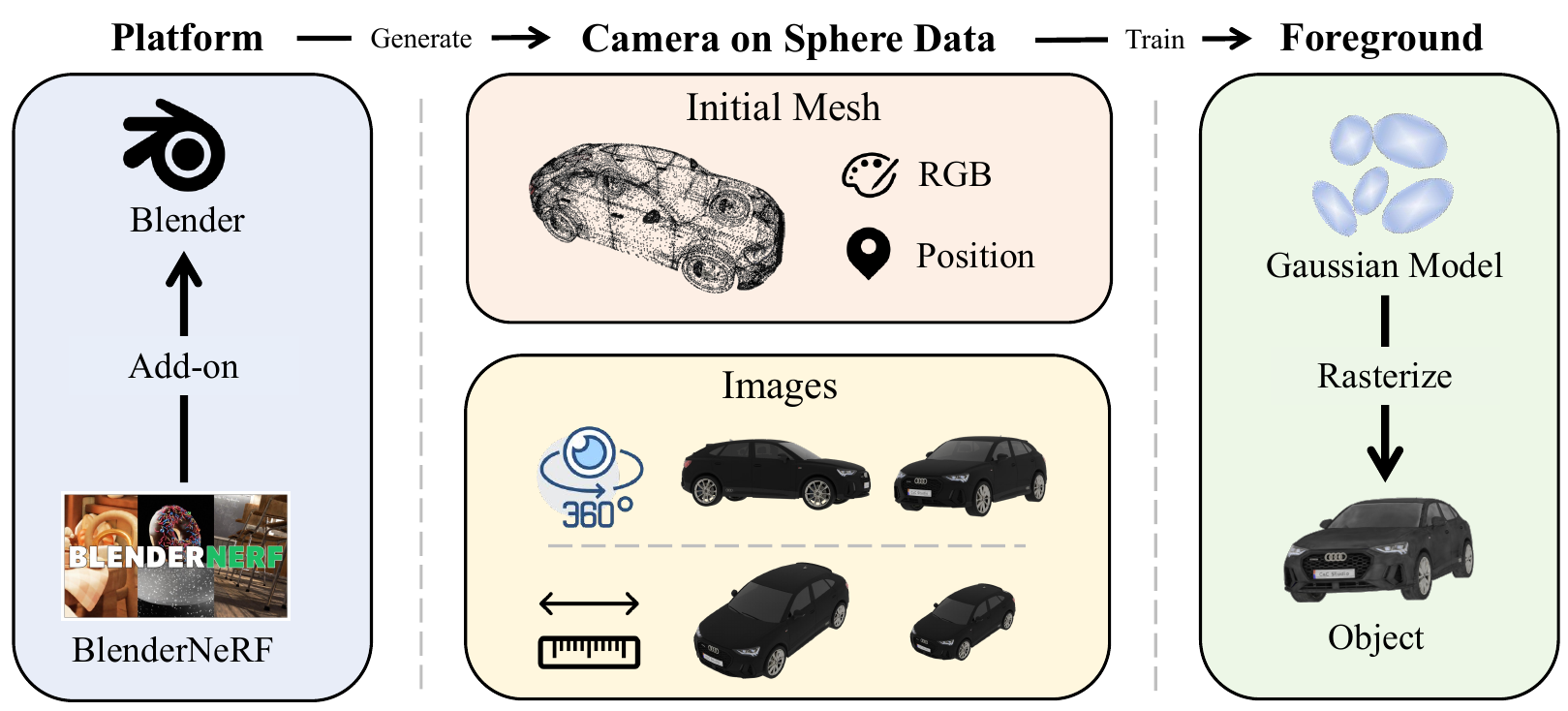}
   \vspace{-2mm}
   \caption{Overview of foreground gaussian model training pipeline for virtual assets. We used Blender with add-on BlenderNeRF to generate camera-on-sphere data, which is collected along a spherical trajectory with a specified radius. Generated data including images and a intial colored mesh is then used to train foreground Gaussian model.}
   \label{fig:foreground_pipeline}
   \vspace{-4mm}
\end{figure*}

\noindent \textbf{Virtual Asset Library Editing and Rendering.}
The pipeline of foreground gaussian model training for each virtual asset is illustrated in Figure \ref{fig:foreground_pipeline}. To ensure that the rendered results of the 3D Gaussian Model align with real-world scene requirements and remove all reflection of asset, we apply the following steps in Blender:\begin{enumerate}
\item \textbf{Creating a Colored Initial Mesh.} We first create a new variable in \textit{Data - Color Attributes - Color}, setting its domain to Vertex and its data type to Color. Then, using \textit{Render - Bake}, we transfer the original material's color to the newly added color variable.

\item \textbf{Enabling Transparent Backgrounds.} To ensure that the rendered images contain only the vehicle and require no additional segmentation before training, we enable the \textit{Render - Film - Transparent} option.

\item \textbf{Adjusting Lighting for Balanced Exposure.} To avoid overly dark or overexposed images, we configure the environment lighting under \textit{World - Surface} by setting the background's HSV values to H=0,S=0,V=0.75.

\item \textbf{Remove Reflection.} We set the roughness of reflective and metallic material to be 1.0. This ensures that the rendering appearance of objects remains consistent across different perspectives, producing identical visual results at the same position regardless of the viewpoint.

\item \textbf{Ensuring Quality Across Varying Distances.} To maintain consistent rendering quality of foreground models at different distances, we use BlenderNeRF\cite{Raafat_BlenderNeRF_2024} to render images of the same 3D asset at 4m, 10m, and 15m distances. For each distance, 150 images are rendered.
\end{enumerate}Following the above setup, for each 3D asset, we generate a total of 450 images captured from different distances and angles, along with the corresponding camera transforms for each image. Besides, we obtain an initial mesh with color attributes. The number of vertices in this mesh depends on the original vehicle model's resolution, ranging from 100,000 to 600,000 vertices.

\noindent \textbf{Virtual Asset Library Model Training.} We use all the rendered images, camera parameters, and the initial mesh from the above process as training data. The initial mesh is downsampled to a maximum of 100,000 vertex to ensure the resulting model remained compact. For each model, the training process follows that of 3D Gaussian Splatting\cite{kerbl20233d}. Specifically, we trained the model for 30,000 iterations.

\noindent \textbf{3DRealCar Library Model Training.} We followed the preprocessing steps provided by 3DRealCar to process the raw images, obtaining the vehicle’s initial mesh and mask. The model training pipeline is consistent with 3D Gaussian Splatting. However, as noted in 3DRealCar~\cite{du20243drealcar}, reflection in the original dataset impact the training quality. To address this issue easily, we set the original image backgrounds to green and purple, alternately using different backgrounds during training. Throughout the process, we progressively removed background color points to obtain the final 3DGS model at every 1,500 iteration. In the future, we plan to explore more efficient methods for rapidly generating clean foreground vehicle models.

\section{Experiment Settings on Runtime Efficiency}
\label{sec:runtime}
\noindent In this section, we validate that \acronym\ achieves real-time rendering capabilities and demonstrates significantly faster performance compared to other simulation platforms that generate high-fidelity images.

All 3D Gaussian Models in \acronym\ are trained on a single NVIDIA GeForce RTX 4090 GPU. For Street Gaussians\cite{yan2024sg} and \acronym\ , we train a scene without any dynamic foreground objects from the Waymo Open Dataset\cite{sun2020scalability}. During rendering, Street Gaussians generate images using the original settings, while \acronym\ includes additional foreground models. Rendering for NeuroNCAP\cite{ljungbergh2025neuroncap} and DriveArena\cite{yang2024drivearena} is performed directly using their provided checkpoints. All rendering experiments are conducted on a single NVIDIA GeForce RTX 4090 GPU. Frame rates are calculated based on the time required to render all views for a single frame. In \acronym\ and NeuroNCAP, we render six 1600×900 images per frame. In DriveArena, a 2400×224 image is rendered, but unlike the original DriveArena setup, we do not further divide it into six 400×224 images or upsample them to a 1600×900 resolution.

\begin{table}[t]
\footnotesize
\begin{center}
\centering
\vspace{-2mm}
\renewcommand{\arraystretch}{1.3}
\setlength{\tabcolsep}{0.3mm}{
\begin{tabular}{ccccc}
\toprule
\textbf{} & \textbf{NeuroNCAP\cite{ljungbergh2025neuroncap}} & \textbf{DriveArena\cite{yang2024drivearena}} & \textbf{Street Gaussians\cite{yan2024sg}} & \textbf{\acronym\ } \\
\midrule \textbf{FPS} & 1.282 & 0.198 & 18.922 & 11.574 \\
\bottomrule
\end{tabular}
}
\vspace{-3mm}
\caption{Quantitative runtime results on different closed-loop simulation platform renderer.}
\label{tab:efficiency}
\vspace{-7mm}
\end{center}
\end{table}

\noindent \textbf{Results.} \acronym\ demonstrates a clear advantage in balancing real-time performance with high-fidelity rendering. While the introduction of additional foreground models and the model fusion process slightly increases computational demand compared to Street Gaussians, \acronym\ effortlessly maintains real-time rendering capabilities. Compared to existing methods based on NeRF and diffusion models, \acronym\ delivers significantly faster rendering speeds along with spatial-temporal consistency. This unique combination of efficiency and quality positions \acronym\ as a potential method for real-time, high-resolution simulation. It not only meets the demands of real-time simulation platforms, reaffirming its superior capability in rendering dynamic, interactive scenarios.

\section{Ego Speed Alteration and Interaction Rate }
\label{sec:interaction_judge}
\subsection{Ego Speed Alteration}
For each planning frame \( t \), we calculate the future trajectory offset between two consecutive future frames, \( \{(w^{t}_{x}, w^{t}_{y}), (w^{t+1}_{x}, w^{t+1}_{y}), \dots, (w^{t+5}_{x}, w^{t+5}_{y})\} \). For each timestamp \( i \) from 0 to 6, we compute the difference between the magnitudes of consecutive waypoints, \( \left\| w^{t+i+1} \right\|_{2} - \left\| w^{t+i} \right\|_{2} \), to determine whether the vehicle is accelerating or decelerating. When the trend switches from acceleration to deceleration, or vice versa, we increment the count for Ego Speed Alteration for the frame \(t\) by 1.

\subsection{Interaction Judgment}
\begin{figure}
  \centering
   \includegraphics[width=\linewidth]{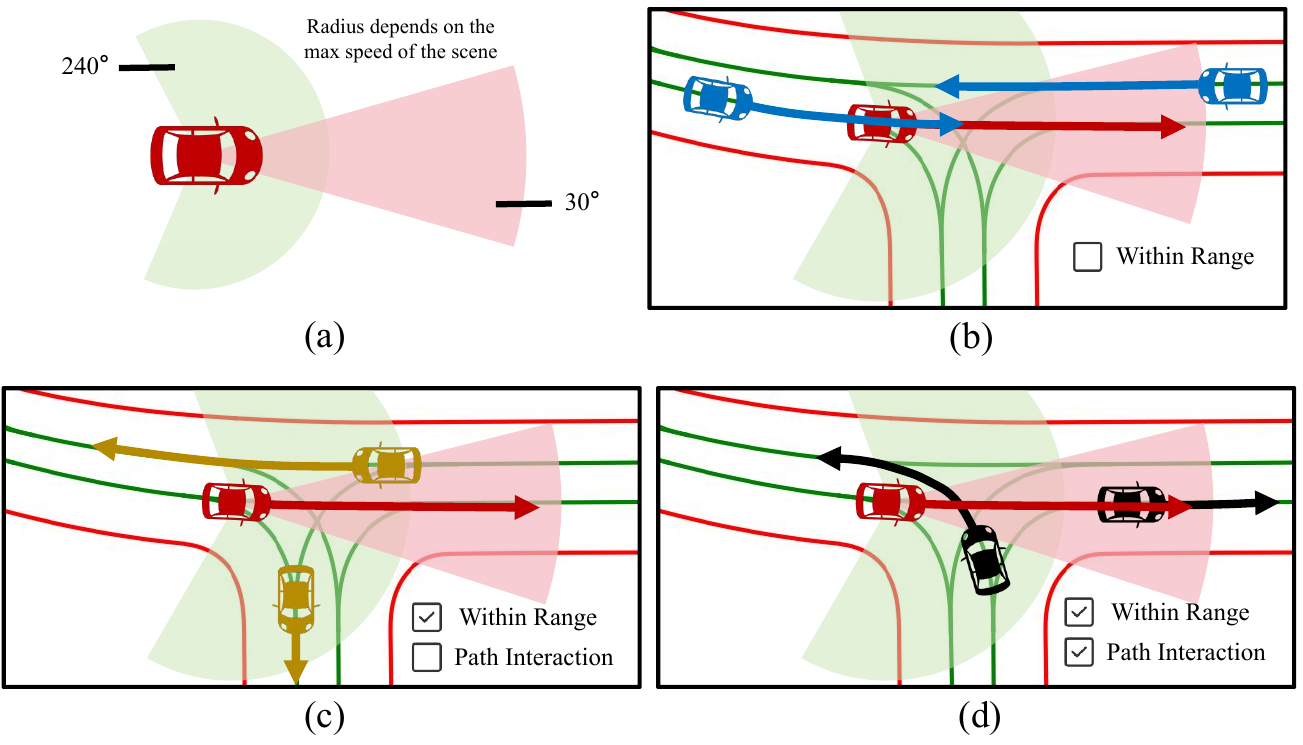}
   \vspace{-5mm}
   \caption{Example of interaction. (a) Illustrates the interaction range, which consists of two parts: a 240° surrounding zone and a 30° forward zone. The radius is determined by the maximum speed of ego vehicle in the original scene data. The vehicles in (b) condition are not considered to be interacting with the ego vehicle as they fall outside the interaction range. Although the vehicles in (c) condition are within the interaction range, their routes do not intersect with ego vehicle's route, which will not be classified as interacting. The vehicles in (d) condition are both within the interaction range and their routes interact with ego vehicle, thus they are considered to be interacting.}
   \vspace{-2mm}
   \label{fig:interaction}
\end{figure}

The determination of interaction consists of two key criteria. First, whether the vehicle is within the defined interaction range. Second, whether its route intersects with that of ego vehicle. A vehicle is only considered to be interacting if both conditions are met, as illustrated in ~\cref{eq:interaction}. Also example are given in ~\cref{fig:interaction}.
\begin{equation}
    \mathbb{I}_{\text{detect}}(v) =
    \begin{cases}
        1, & \text{if } v \text{ is within range} \\
        0, & \text{otherwise}
    \end{cases}
\end{equation}

\begin{equation}
    \mathbb{I}_{\text{interact}}(v) =
    \begin{cases}
        1, & \text{if path of } v \text{ intersects} \\
        0, & \text{otherwise}
    \end{cases}
\end{equation}

\begin{equation}
    \mathbb{I}_{\text{interaction}}(v) = \mathbb{I}_{\text{detect}}(v) \cdot \mathbb{I}_{\text{interact}}(v)
    \label{eq:interaction}
\end{equation}

\section{Justification for comparison with other mainstream simulator}
\label{sec:justify}
We do not compare our experimental results with other driving simulators such as CARLA~\cite{dosovitskiy2017carla}, DriveArena~\cite{yang2024drivearena}, and HugSim~\cite{zhou2024hugsim} due to fundamental differences in their design and capabilities. CARLA lacks sufficient realism in sensor simulation and vehicle dynamics, making it unsuitable as input for end-to-end model trained on real world data. DriveArena is designed around the nuScenes dataset, and adapting it to the Waymo dataset would require significant effort and computational resources, making a fair comparison impractical. HugSim, on the other hand, does not incorporate an expert planner and is unable to generate comprehensive driving logs that include both sensor inputs and ego dynamics. Given these limitations, a direct experimental comparison with these simulators would not provide meaningful insights.

\section{Spatial-temporal Consistency}
\label{sec:spa-temp-consist}
\begin{figure*}
  \centering
   \includegraphics[width=\linewidth]{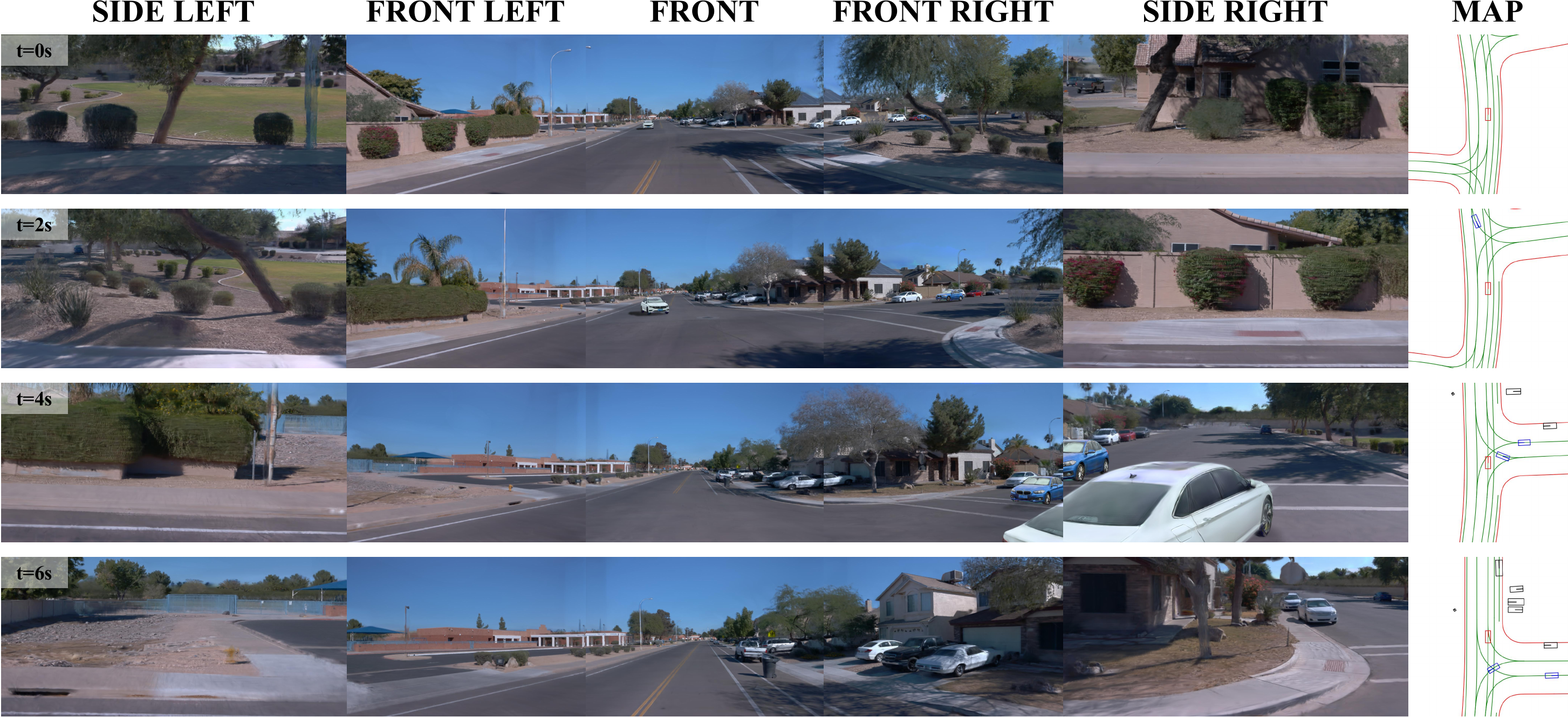}
   \vspace{-2mm}
   \caption{Reconstructed perspectives from multi-view cameras across sequential time steps. Each row represents a specific timestamp, with time advancing incrementally downward. The visualization demonstrates the rendering outcomes of surround-view cameras over time, highlighting spatial-temporal consistency of \acronym\ . The original video is also in the supplementary material.}
   \label{fig:consistency}
   \vspace{-4mm}
\end{figure*}

As mentioned in Section \ref{scene_renderer}, the parameters of the composite Gaussian model are fully defined for each frame, with only the vehicle positions changing over time. Consequently, \acronym\ is inherently capable of maintaining spatial-temporal consistency during the rendering process. As shown in Figure \ref{fig:consistency}, both the background and the incorporated foreground objects consistently retain their coherence across consecutive renderings.

\section{Closed-loop Evaluation Metrics}
\label{sec:metrics}
Our closed-loop evaluation follows CARLA, calculating key metrics such as route completion (RC), vehicle collision rate (VCR), and layout collision rate (LCR). 
Vehicle collision rate (VCR) is calculated as:
\begin{equation}
VCR = \frac{1}{N} \sum_{i=1}^{N} \frac{1}{T_i} \sum_{j=1}^{T_i} Col_{vehicle}(j)
\end{equation}

where \(N\) is the number of scenarios, \(T_i\) is the number of the \(i\)-th scenario frames, and \(Col_{vehicle}(j)\) indicates whether ego vehicle collides with other vehicles in \(j\)-th frame.

Similarly, layout Collision Rate (LCR) is calculated as:
\begin{equation}
LCR = \frac{1}{N} \sum_{i=1}^{N} \frac{1}{T_i} \sum_{j=1}^{T_i} Col_{layout}(j)
\end{equation}

The Route Completion (RC) is determined based on the absence of vehicle collisions and timeouts, as follows:
\begin{equation}
RC = \frac{1}{N} \sum_{i=1}^{N} \delta(VCR_i = 0 \land \neg \text{timeout}_i)
\end{equation}
where \(\text{timeout}_i\) indicates whether ego vehicle’s behavior timed out in the \(i\)-th scenario, and the function \(\delta(\cdot)\) returns 1 when the condition is met, and 0 otherwise. The calculation of RC excludes LCR because we observe that the original model frequently collides with the layout in our created scenarios, resulting in zero RC score across the board, which diminishes its comparative value.

\section{Adaptive Kinematic Model Estimation}
Given real world IMU data from two consecutive frames $(x_t,y_t,\varphi_{t},v_t,x_{t+1},y_{t+1},\varphi_{t+1},v_{t+1})$ in waymo with $\Delta t=0.1s$, we can obtain based on the second line of ~\cref{eq:AKM} that:
\[
\beta_t = \arcsin\left(\frac{l_f}{v_t \Delta t} (\varphi_{t+1} - \varphi_t) \right)
\]  
where $\beta_t$ is the slip angle, $v_t$ is the vehicle velocity, $\varphi_t$ is the yaw angle and $x_t$, $y_t $ represent the vehicle position.
Then we aim to obtain the best value of $u_1$ and $u_2$ by solving the following optimization problem:
\[
\begin{aligned}
    \min_{u_1, u_2} \quad & \sum^{N}_{n=1}\sum^{T_n}_{t=1} \left( (x_{t+1} - \hat{x}_{t+1})^2 + (y_{t+1} - \hat{y}_{t+1})^2 \right) \\
    \text{subject to:} \quad & \\
    v_u &= (1 - \boldsymbol{u_1}) v_t + \boldsymbol{u_1} v_{t+1}, \\
    \hat{x}_{t+1} &= x_t + v_u\cos(\varphi_t + \boldsymbol{u_2} \cdot \beta_t) \Delta t, \\
    \hat{y}_{t+1} &= y_t + v_u\sin(\varphi_t + \boldsymbol{u_2} \cdot \beta_t) \Delta t.
\end{aligned}
\]
Where $N$ represent the number of scenes and $T_n$ represent the number of frame pairs in the $n$ th scene.

\section{End-to-End Model Training Details}
The training of the E2E model follows the official repositories of VAD and GenAD. To align with the Waymo Open Dataset settings, we use five cameras and set the output dimension of the detection classification head to four, corresponding to the four categories in the dataset. Training is stopped at epoch 30, as the validation loss begins to increase beyond this point. For fine-tuning on real data, we use the same configuration as the initial training while adjusting the initial learning rate to $lr=5e^{-5}$

\section{Additional Closed-Loop Results}
~\cref{tab:close_loop1} presents the performance of original and fine-tuned models on scenes from $\mathcal{S}_{recon}$. Surprisingly, the original model performs poorly in closed-loop evaluation, despite being trained on real-world logs from the same scenes. We attribute this to a distributional bias in the training set, where approximately one-third of samples involve only acceleration or deceleration without phase changes. As a result, imitation-learning-based E2E models tend to adopt a flawed pattern—either continuously accelerating until a collision with other agents or decelerating to a stop. Our results show that synthetic data helps mitigate this bias, leading model to learn more reasonable driving patterns.

\begin{table}[H]
\vspace{-1mm}
\footnotesize
\begin{center}
\centering
\setlength{\tabcolsep}{1.9mm}{
\begin{tabular}{lcccccc}
\toprule
\multirow{2}{*}{\textbf{Method}} & \multicolumn{3}{c}{\textbf{All Test Scenes}} & \multicolumn{3}{c}{\textbf{Turn Scenes}} \\
\cmidrule(lr){2-4} \cmidrule(lr){5-7}
 & \textbf{RC$\uparrow$} & \textbf{VC$\downarrow$} & \textbf{LCR$\downarrow$} & \textbf{RC$\uparrow$} & \textbf{VC$\downarrow$} & \textbf{LCR$\downarrow$} \\
\midrule
  \textbf{VAD} & 29.88 & 8.35 & 8.62 & 19.44 & 7.92 & 11.67 \\
  \textbf{VAD\textsubscript{ft}} & \textbf{39.02} & \textbf{6.93} & \textbf{5.67} & \textbf{31.15} & \textbf{5.71} & \textbf{4.53} \\
\bottomrule
\end{tabular}
}
\vspace{-3mm}
\caption{Closed-Loop Evaluation of different models across $S_{recon}$. Performance metrics include: \textbf{RC} (Route Completion), \textbf{VC} (Vehicle Collision Rate, and \textbf{LCR} (Layout Collision Rate).}
\vspace{-10mm}
\label{tab:close_loop1}
\end{center}
\end{table}

\begin{table*}[t]
\begin{center}
\centering
\vspace{-5mm}
\begin{tabular}{cccc}
\hline
Index & Segment Name                                                             & Time  & Type     \\ \hline
019   & segment-10275144660749673822\_5755\_561\_5775\_561\_with\_camera\_labels & Day   & Turn     \\
036   & segment-10676267326664322837\_311\_180\_331\_180\_with\_camera\_labels   & Day   & Straight \\
053   & segment-11017034898130016754\_697\_830\_717\_830\_with\_camera\_labels   & Day   & Straight \\
078   & segment-11623618970700582562\_2840\_367\_2860\_367\_with\_camera\_labels & Day   & Turn     \\
155   & segment-13390791323468600062\_6718\_570\_6738\_570\_with\_camera\_labels & Day   & Turn     \\
250   & segment-15445436653637630344\_3957\_561\_3977\_561\_with\_camera\_labels & Day   & Turn     \\
276   & segment-15943938987133888575\_2767\_300\_2787\_300\_with\_camera\_labels & Day   & Turn     \\
367   & segment-17818548625922145895\_1372\_430\_1392\_430\_with\_camera\_labels & Day   & Turn     \\
382   & segment-18111897798871103675\_320\_000\_340\_000\_with\_camera\_labels   & Day   & Turn     \\
402   & segment-1918764220984209654\_5680\_000\_5700\_000\_with\_camera\_labels  & Day   & Straight \\
427   & segment-2259324582958830057\_3767\_030\_3787\_030\_with\_camera\_labels  & Day   & Turn     \\
451   & segment-268278198029493143\_1400\_000\_1420\_000\_with\_camera\_labels   & Night & Straight \\
619   & segment-580580436928611523\_792\_500\_812\_500\_with\_camera\_labels     & Night & Straight \\
674   & segment-7007702792982559244\_4400\_000\_4420\_000\_with\_camera\_labels  & Night & Straight \\
739   & segment-8513241054672631743\_115\_960\_135\_960\_with\_camera\_labels    & Night & Straight \\ \cline{1-3}
\end{tabular}
\vspace{-1mm}
\caption{Information for some selected scenarios in the Waymo Open Dataset.}
\label{tab:information_of_data}
\vspace{-5mm}
\end{center}
\end{table*}

\begin{figure*}
  \centering
   \includegraphics[width=0.9\linewidth]{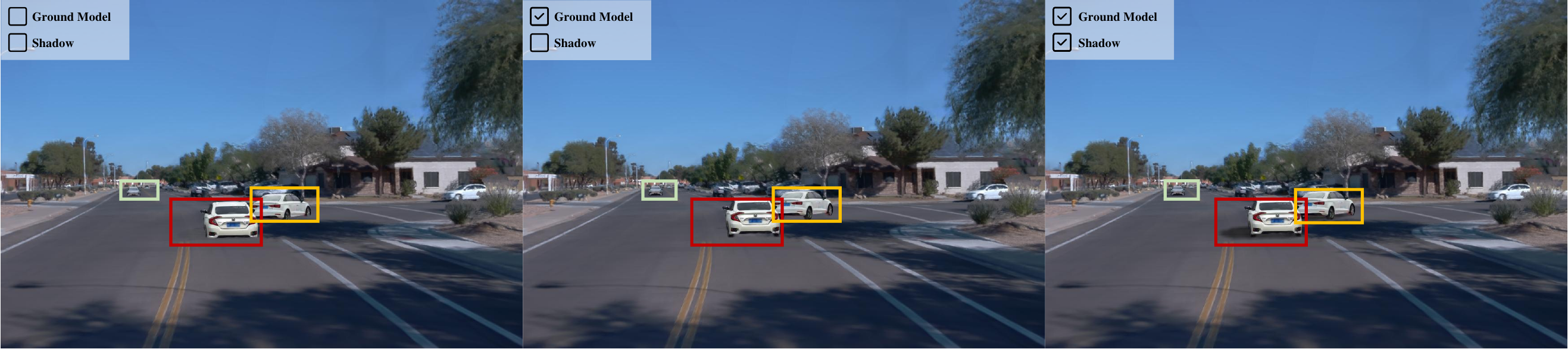}
   \vspace{-2mm}
   \caption{Example of the effect of ground model and shadow addition.}
   \label{fig:ground_comparison}
   \vspace{-4mm}
\end{figure*}


\end{document}